%% file: main.tex
\pdfoutput=1
\PassOptionsToPackage{table}{xcolor}
\documentclass{article}
\usepackage{style/fancyhdr}
\usepackage{style/natbib}
\makeatletter
\@namedef{ver@fancyhdr.sty}{}
\@namedef{ver@natbib.sty}{}
\makeatother
\usepackage{style/iclr2027_conference}
\usepackage{times}

\usepackage{amsmath}
\usepackage{amssymb}
\usepackage{array}
\usepackage{booktabs}
\usepackage{float}
\usepackage{graphicx}
\usepackage{longtable}
\usepackage{multirow}
\usepackage{needspace}
\usepackage{placeins}
\usepackage{tabularx}
\usepackage{wrapfig}
\usepackage{xcolor}
\usepackage{tikz}

\definecolor{linkblue}{rgb}{0.21,0.49,0.74}
\usepackage[colorlinks=true,allcolors=linkblue]{hyperref}
\hypersetup{pdftitle={HumanSkill-4D: Training-Free Language Querying of Animatable Gaussian Avatars in Space and Time},pdfauthor={Junsu Kim, Naeun Kim, Dongyoon Han, Seungryul Baek}}

\definecolor{wherered}{HTML}{D81B60}
\definecolor{whenblue}{HTML}{00883C}
\definecolor{offlineblue}{HTML}{2E5FAE}
\definecolor{onlinepurple}{HTML}{5B45B8}
\definecolor{oursrow}{HTML}{E6EEF9}
\definecolor{grouprow}{HTML}{ECEFF2}
\definecolor{unistnavy}{HTML}{2F5FC4}
\definecolor{navergreen}{HTML}{03C75A}
\newcommand{\RefThreeD}{ZJU-Refer-3D}
\newcommand{\RefHuMManFourD}{HuMMan-Refer-4D}
\newcommand{\RefZJUFourD}{ZJU-Refer-4D}
\newcommand{\Ours}{\mbox{HumanSkill-4D}}
\title{\Ours{}: Training-Free Language\\Querying of Animatable Gaussian Avatars\\in Space and Time}
\author{\makebox[\dimexpr\textwidth-2\tabcolsep\relax][c]{Junsu Kim$^{\textcolor{unistnavy}{\spadesuit}}$ \quad Naeun Kim$^{\textcolor{unistnavy}{\spadesuit}}$ \quad Dongyoon Han$^{\textcolor{navergreen}{\clubsuit},\dagger}$ \quad Seungryul Baek$^{\textcolor{unistnavy}{\spadesuit},\dagger}$} \\
\makebox[\dimexpr\textwidth-2\tabcolsep\relax][c]{\normalfont $^{\textcolor{unistnavy}{\spadesuit}}$UNIST \quad $^{\textcolor{navergreen}{\clubsuit}}$NAVER AI Lab \quad $^{\dagger}$Corresponding authors}}

\iclrfinalcopy

\begin{document}
\maketitle
\lhead{Preprint}

\begin{figure}[H]
\vspace{-26pt}
\centering
\includegraphics[width=.90\linewidth]{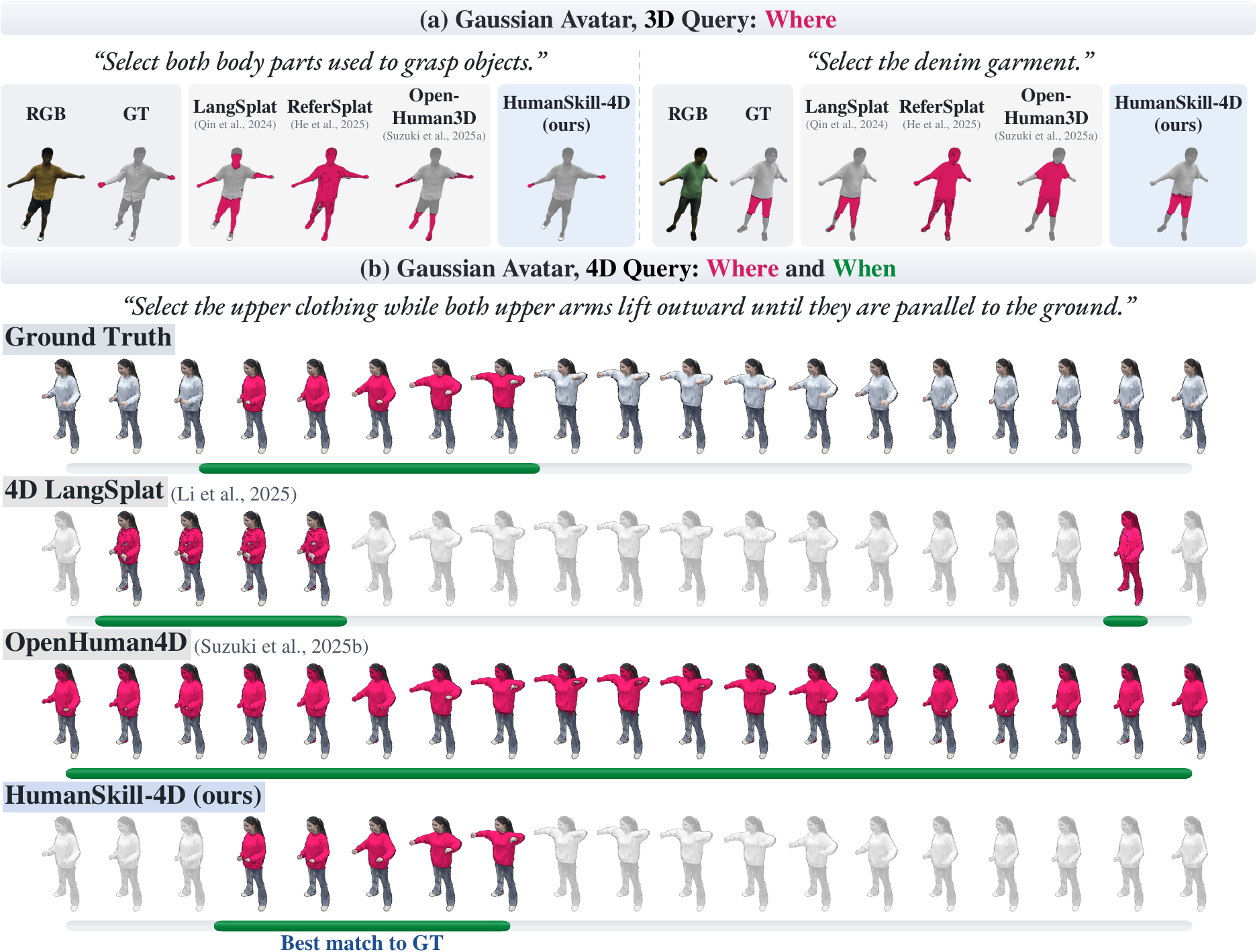}
\vspace{-0.75em}
\caption{\textbf{Avatar referring segmentation with \Ours{}.} \textbf{(a)} A 3D query selects the described region of a Gaussian avatar (\textcolor{wherered}{\emph{Where}}). \textbf{(b)} A 4D query also selects the intervals in which the specified pose or motion occurs (\textcolor{whenblue}{\emph{When}}). See our \href{https://jjunsss.github.io/HumanSkill-4D/}{project page} for video demos.}
\label{fig:queries}
\vspace{-1.5em}
\end{figure}

\begin{abstract}
We introduce avatar referring segmentation, which selects regions of animatable Gaussian avatars from language queries. A query can also specify a pose or motion to select matching intervals. This lets users interact with avatars for local editing and motion inspection. However, making these selections with existing language Gaussian splatting methods poses two challenges: adjacent parts and garments are hard to separate and may be referred to indirectly (\emph{region ambiguity}), while fine-grained motions can look alike in RGB (\emph{motion ambiguity}). We address these challenges with \Ours{}, a training-free framework that shares a multimodal large language model (MLLM) and geometric operators across avatars. For each avatar, a semantic memory links body-part and garment categories to persistent Gaussian regions and appearance descriptions. The MLLM uses category names and descriptions to select categories by appearance, relation, or function. To find matching intervals, it composes operators into programs that check query conditions on driving joint trajectories stored in a motion memory. These category choices and operator outputs make selections traceable, while composing or adding operators supports new motion conditions. To evaluate this task, we introduce three human-annotated benchmarks and show that \Ours{} outperforms the baselines in part segmentation and 3D and 4D referring segmentation, with compact memories and only minutes of setup per avatar.
\end{abstract}

\section{Introduction}
\label{sec:intro}
\vspace{-1em}

Gaussian avatars reconstruct a person's geometry and appearance and animate the result using a driving pose sequence~\citep{qian2024gsavatar,kocabas2024hugs}. In applications such as AR/VR and sports~\citep{hu2024gauhuman,jiang2025dnfavatar}, users may need to edit an avatar's appearance or inspect its motion. Clothing transfer and appearance editing~\citep{lin2024layga,ho2023locallyeditable}, for example, require selecting the body part or garment to modify (\textcolor{wherered}{\emph{Where}}). Inspecting that region during a particular pose or motion also requires finding the intervals in which that pose or motion occurs (\textcolor{whenblue}{\emph{When}}). A language query can describe both, but the avatar has no part or garment labels on its Gaussians and no motion labels on its frames. We therefore introduce avatar referring segmentation: a 3D query selects the described region (Fig.~\ref{fig:queries}(a)), while a 4D query also selects the intervals that match the specified pose or motion (Fig.~\ref{fig:queries}(b)).

In reconstructed scenes, language Gaussian splatting already selects regions and relevant time intervals. In 3D, these methods attach image-language features to Gaussians and select regions whose features match a query~\citep{qiu2024featuresplatting}. These features are learned per scene as a language field in LangSplat~\citep{qin2024langsplat} or registered directly in Dr.~Splat~\citep{kim2025drsplat}, whereas ReferSplat~\citep{he2025refersplat} learns per-scene referring features from annotated expressions. To also determine when a query is relevant, 4-LEGS~\citep{fiebelman2025_4legs} grounds video-text features in dynamic Gaussians, while 4D LangSplat~\citep{li2025_4dlangsplat} learns a temporal language field from captions generated by a multimodal large language model (MLLM). For humans, OpenHuman3D~\citep{suzuki2025openvocab3dhuman} segments body parts and garments with a CLIP image encoder fine-tuned on human parsing data, and OpenHuman4D~\citep{suzuki2025openhuman4d} tracks part masks over time.

Despite these advances, enabling language querying on Gaussian avatars presents two challenges. The first is \emph{region ambiguity}. Human parts and garments lie close together and often overlap. As the pose or viewpoint changes, different boundaries become visible. Feature-matching methods~\citep{qin2024langsplat} rely on class-agnostic SAM masks~\citep{kirillov2023sam}, which do not necessarily follow these boundaries, and on generic CLIP~\citep{radford2021clip} features, which poorly distinguish body parts and garments~\citep{suzuki2025openvocab3dhuman}. Replacing the SAM masks with human-parser masks improves the overall accuracy of LangSplat and Dr.~Splat, but leaves their CLIP-based matching unchanged. Queries may also refer to a part or garment indirectly, through its relation to another region or its function, and CLIP-based matching remains unreliable for such references in our evaluation (Tab.~\ref{tab:static3d}). The second is \emph{motion ambiguity}. Even after the target region is identified, locating the intervals of a fine-grained pose or motion remains difficult (e.g., \emph{``Select the upper clothing while both upper arms lift outward until they are parallel to the ground''} in Fig.~\ref{fig:queries}(b)). Mask tracking~\citep{suzuki2025openhuman4d} keeps the clothing selected over time but does not determine when the arms meet this condition. RGB-based methods can localize intervals~\citep{fiebelman2025_4legs,li2025_4dlangsplat}, but different motions can look alike in RGB frames while differing in their dynamics~\citep{yang2026movie}, and even video MLLMs struggle with fine-grained motion understanding~\citep{hong2025motionbench}.

Gaussian avatars, however, already provide structure suited to both challenges. Each Gaussian is defined in a canonical space and deformed to every pose~\citep{qian2024gsavatar}, so it keeps its identity as the avatar moves. The driving pose sequence also provides joint trajectories. Building on this structure, we introduce \Ours{}, which organizes avatar-specific knowledge and shared procedures into an agent skill~\citep{tang2026wikiskill} for a frozen MLLM. The skill's knowledge lies in per-avatar memories of the Gaussians in each part or garment and of joint trajectories. Its procedures are geometric operators that check pose and motion conditions and are shared across avatars.

Specifically, to address \emph{region ambiguity}, we accumulate labels from a 2D human parser~\citep{khirodkar2026sapiens2} across views and poses into category scores that link body parts and garments to the avatar's Gaussians. A semantic memory combines these scores with category names and MLLM-generated appearance descriptions. Given a query, the MLLM uses the names and descriptions to select categories by appearance, relation, or function, rather than matching the query to CLIP features, and the Gaussians linked to these categories form the selected region. For \emph{motion ambiguity}, we store the driving joint trajectories in a motion memory and check query conditions on them rather than on RGB frames. The checks use geometric operators adapted from PoseScript~\citep{delmas2022posescript}, which describes poses by relations between joints, and MotionScript~\citep{yazdian2025motionscript}, which extends these relations to motion. Given a 4D query, the MLLM composes these operators into a program by linking them with temporal relations, and running it on the trajectories gives the matching intervals. In Fig.~\ref{fig:queries}(b), for example, the program keeps the outward movements of both upper arms that end parallel to the ground. Because the MLLM chooses categories and operators rather than regions or intervals, each selection can be traced to these choices, and new motion conditions only require composing existing operators and relations or adding new ones. To evaluate this task, we introduce three human-annotated benchmarks for selecting regions and pose or motion intervals. \Ours{} performs best in part segmentation and 3D and 4D referring segmentation.

We make four contributions: (1)~we introduce avatar referring segmentation and \Ours{}, a training-free framework that gives the structure of Gaussian avatars to a frozen MLLM as a skill, making its decisions traceable; (2)~to address \emph{region ambiguity}, we build a semantic memory that also supports part segmentation; (3)~to address \emph{motion ambiguity}, we localize motions on joint trajectories with MLLM-composed programs, extensible by composing or adding operators and relations; and (4)~to evaluate this task, we introduce three benchmarks and show that \Ours{} is the most accurate on all three, with compact memories and only minutes of setup per avatar.

\section{Related Work}
\label{sec:related}

\textbf{3D language Gaussian splatting.} Most language Gaussian methods select regions by matching a query to CLIP features that are learned per scene~\citep{qin2024langsplat,shi2024legaussians,wu2024opengaussian} or registered without per-scene optimization~\citep{kim2025drsplat,ding2026extrinsplat}. Beyond object names, ReferSplat~\citep{he2025refersplat} learns per-scene referring features from annotated expressions, and ReLaGS~\citep{xie2026relags} adds relation-guided retrieval. Open-vocabulary part segmentation likewise matches text to language-aligned features of 3D shapes~\citep{ma2025find3d} and humans~\citep{suzuki2025openvocab3dhuman}. Whereas these methods match queries to language features, \Ours{} uses an MLLM to interpret region queries through a semantic memory of avatar parts and garments, supporting both referring and part segmentation.

\textbf{4D language Gaussian splatting.} Recent work extends language Gaussian splatting to dynamic scenes, where a query specifies a region and when it is relevant. 4D LangSplat~\citep{li2025_4dlangsplat} and 4-LEGS~\citep{fiebelman2025_4legs} learn temporal language features from MLLM-generated captions and video-text features, respectively, and ST4R-Splat~\citep{meng2026st4rsplat} extends referring segmentation to 4D Gaussians. For humans, OpenHuman4D~\citep{suzuki2025openhuman4d} and 4D-DRESS~\citep{wang20244d} parse human parts over time. However, these methods mainly rely on RGB evidence, in which different motions can look alike~\citep{yang2026movie} and fine-grained motion remains hard to capture~\citep{hong2025motionbench}. \Ours{} instead measures pose and motion conditions on the driving joint trajectories, so it can distinguish subtle differences, such as the direction and extent of a movement, and return the queried region with its matching intervals using a compact memory.

\textbf{Agent skills and executable procedures.} Large language models increasingly solve complex or fine-grained tasks through supplied procedures rather than task-specific training. Some compose predefined actions or handlers~\citep{khot2023decomposed,pan2025chainofaction}, whereas others generate code over visual tools~\citep{gupta2023visprog,suris2023vipergpt}. More recently, agent skills compile experience into reusable knowledge and procedures that accompany the query~\citep{wang2024voyager,tang2026wikiskill}, and this paradigm now extends to 3D spatial and 4D spatiotemporal reasoning~\citep{li2026skill3d,sohn2026r4}. Building on this paradigm, \Ours{} gives an MLLM per-avatar memories as knowledge and shared geometric operators and temporal relations as procedures, so a new avatar needs only its memories, and a new motion condition requires only composing existing procedures or adding new ones.

\section{Preliminaries}
\label{sec:preliminaries}

\textbf{Gaussian avatars and motion.} 3DGS-Avatar~\citep{qian2024gsavatar} represents an avatar as a fixed set of $N$ Gaussians $g_i$ with position, shape, opacity, and color. These Gaussians are defined in a canonical space corresponding to the reference pose of SMPL~\citep{loper2015smpl}, a body model parameterized by shape and joint rotations. A driving pose sequence of $T$ frames specifies the joint rotations and root translation at each frame, which determine the 3D joint positions. Animation deforms the Gaussians with the body into each target pose, so $g_i$ is the same Gaussian across frames.

\textbf{Rendering.} Rendering projects the Gaussians posed at frame $t$ into camera view $v$ and blends their colors at pixel $u$ from front to back~\citep{kerbl2023gaussian}. Let $\alpha_i^{v,t}(u)$ be the projected opacity of Gaussian $g_i$ at that pixel, and write $j\prec i$ when Gaussian $g_j$ lies in front of $g_i$ along the pixel's ray. The rendering contribution of $g_i$ to pixel $u$, the coefficient of its color in the blended pixel, is
\begin{equation}
w_i^{v,t}(u)=\alpha_i^{v,t}(u)\prod_{j\prec i}\bigl(1-\alpha_j^{v,t}(u)\bigr),
\label{eq:gaussian_rendering}
\end{equation}
where the product is the transmittance through the Gaussians in front of $g_i$.

\section{Method}
\label{sec:method}

\subsection{Overview}
\label{sec:method_overview}

\begin{figure}[t]
\centering
\includegraphics[width=\linewidth]{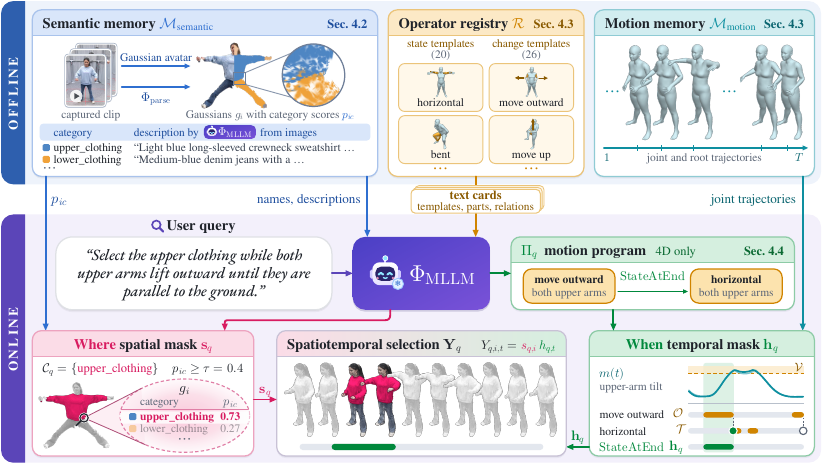}\par
\vspace{-.5em}
\caption{\textbf{\Ours{} overview.} \textbf{\textcolor{offlineblue}{\mbox{(Offline)}}} We build a semantic memory for each avatar, a motion memory for its driving pose sequence, and an operator registry shared across avatars. \textbf{\textcolor{onlinepurple}{\mbox{(Online)}}} The frozen MLLM selects target categories from their names and descriptions, and their stored scores determine the spatial mask. For a 4D query, it also reads the registry's text cards and composes operators into a motion program, and running the program on the joint trajectories gives the temporal mask. The two masks give the spatiotemporal selection. In this example, the program identifies intervals when both upper arms move outward, and \textit{StateAtEnd} retains each movement interval whose last frame has both arms horizontal (orange band).}
\label{fig:pipeline}
\vspace{-6pt}
\end{figure}

Given a language query, \Ours{} selects the described region of an avatar and, for a 4D query, every matching interval in its driving pose sequence. As shown in Fig.~\ref{fig:pipeline}, the framework prepares reusable memories offline, then uses a frozen MLLM $\Phi_{\mathrm{MLLM}}$ to interpret queries online. 

In the offline phase, we build a semantic memory $\mathcal{M}_{\text{semantic}}$ that links body-part and garment categories to the avatar's Gaussians through category scores $p_{ic}$ registered from a frozen 2D human parser $\Phi_{\mathrm{parse}}$. The memory also stores category-level appearance descriptions generated by $\Phi_{\mathrm{MLLM}}$ from multi-view images (Sec.~\ref{sec:method_memory}). A motion memory $\mathcal{M}_{\text{motion}}$ stores joint and root trajectories for each driving pose sequence. A registry $\mathcal{R}$ shared across avatars provides reusable templates for checking pose and motion conditions, along with temporal relations that specify how these conditions overlap or follow one another in time (Sec.~\ref{sec:method_operators}).

In the online phase, $\Phi_{\mathrm{MLLM}}$ reads the query $q$ and the category names and descriptions to select target categories $\mathcal{C}_q$. Their stored scores determine the spatial mask $\mathbf{s}_q$ (\textcolor{wherered}{\emph{Where}}), which is the complete output for a 3D query. For a 4D query, the model also reads the registry's text cards and composes operators into a motion program $\Pi_q$. A deterministic executor runs this program on $\mathcal{M}_{\text{motion}}$ to obtain the temporal mask $\mathbf{h}_q$ (\textcolor{whenblue}{\emph{When}}). The two masks give the spatiotemporal selection $\mathbf{Y}_q$ (Sec.~\ref{sec:method_inference}).

\subsection{Building the Semantic Memory}
\label{sec:method_memory}

Given a Gaussian avatar reconstructed with 3DGS-Avatar~\citep{qian2024gsavatar}, we retain its canonical Gaussians $g_i$. From the captured images, we select a registration set $\mathcal{D}_{\text{reg}}$ of camera-frame pairs. For each $(v,t)\in\mathcal{D}_{\text{reg}}$, a frozen 2D human parser $\Phi_{\mathrm{parse}}$ labels each person pixel $u$, and we map this label to $\ell^{v,t}(u)\in\mathcal{K}$, where $\mathcal{K}$ contains the body-part and garment categories (details in Supp.~Sec.~\ref{app:categories}).

Because each Gaussian retains its identity across poses, we aggregate labels across views and frames using contribution-based Gaussian registration inspired by \citet{kim2025drsplat}. For each $(v,t)$, we render the posed avatar through camera $v$ and weight each pixel's label by the Gaussian's rendering contribution $w_i^{v,t}(u)$ from Eq.~\ref{eq:gaussian_rendering}. These weighted labels give the category score
\begin{equation}
    p_{ic}=\frac{\sum_{(v,t)\in\mathcal{D}_{\text{reg}}}\sum_u w_i^{v,t}(u)\,\mathbf{1}[\ell^{v,t}(u)=c]}{\sum_{(v,t)\in\mathcal{D}_{\text{reg}}}\sum_u w_i^{v,t}(u)\,\mathbf{1}[\ell^{v,t}(u)\in\mathcal{K}]}.
\label{eq:semantic_registration}
\end{equation}
Here, $\mathbf{1}[\cdot]$ denotes the indicator function. The score $p_{ic}$ is the share of category $c$ in Gaussian $g_i$'s rendering contribution to pixels with a label in $\mathcal{K}$. If the denominator is zero, we set $p_{ic}=0$ for all $c$. We retain all category scores and associate $g_i$ with every category whose score is at least $\tau$.

To support appearance-based references such as \emph{``the denim garment''} in Fig.~\ref{fig:queries}(a), we augment category names with avatar-specific descriptions. The frozen $\Phi_{\mathrm{MLLM}}$ receives the category names and four camera views taken at the frame closest to a T-pose. It summarizes visible color, texture, graphics, and garment coverage into one description per category (instruction in Supp.~Tab.~\ref{tab:prompt_describer}). Together, the category names, descriptions, and category scores form the reusable per-avatar semantic memory $\mathcal{M}_{\text{semantic}}$ used for online region selection.
\subsection{Building the Motion Memory and Operator Registry}
\label{sec:method_operators}

\textbf{Motion memory.} The motion memory $\mathcal{M}_{\text{motion}}$ stores SMPL joint positions, root orientation, and root translation for all $T$ frames of the driving pose sequence. For directional measurements, we adapt MotionScript's first-frame normalization~\citep{yazdian2025motionscript} to define a coordinate frame fixed throughout the sequence. The initial root position defines its origin, gravity defines its vertical axis, and the person's initial facing direction defines its forward axis.

\textbf{Operator registry.} The operator registry $\mathcal{R}$ is shared across avatars and stores templates for checking pose and motion conditions, along with temporal relations that specify how these conditions overlap or follow one another in time. A template is a reusable rule that specifies a geometric measurement and a condition to check, without selecting a particular body part. For example, the \emph{horizontal} template in Fig.~\ref{fig:pipeline} measures a segment's tilt relative to gravity and checks whether it lies in the range for a horizontal pose. Each template also specifies its coordinate frame and compatible body parts. State templates check conditions at individual frames, whereas change templates check how measurements change over intervals. We adapt these measurements and conditions from PoseScript~\citep{delmas2022posescript} and MotionScript~\citep{yazdian2025motionscript}, defining 46 templates---20 state and 26 change templates---in seven measurement types: joint angle, relative position, displacement, root rotation, segment tilt, distance, and height above the lowest joint (Supp.~Sec.~\ref{app:registry}). The registry also provides a text card for each template and relation, which describes it in words and is later given to $\Phi_{\mathrm{MLLM}}$ with the query.

\textbf{From templates to operators.} Applying a template to a compatible body part defines an executable operator $\rho$. The selected body part determines which SMPL joint trajectories the operator reads from $\mathcal{M}_{\text{motion}}$. For example, applying the \emph{horizontal} template to the left upper arm selects the left shoulder--elbow segment. In Fig.~\ref{fig:pipeline}, applying the \emph{move outward} and \emph{horizontal} templates to both upper arms gives two outward-motion operators $\rho_1,\rho_2$ and two horizontal-state operators $\rho_3,\rho_4$. Let $m_\rho(t)$ denote the measurement of operator $\rho$ at frame $t$. A state operator selects frames where $m_\rho(t)$ lies in the template's valid range $\mathcal{V}_\rho$. A change operator first identifies candidate frame intervals $[a,b]\in\mathcal{O}_\rho^{\mathrm{cand}}$ over which the measurement changes in the specified direction, then retains those meeting a minimum total change $\Delta_\rho^{\min}$ and a minimum interval length $L_{\min}$ (details in Supp.~Sec.~\ref{app:floors}). The resulting state frames $\mathcal{T}_\rho$ and change occurrences $\mathcal{O}_\rho$ are
\begin{equation}
\begin{aligned}
\mathcal{T}_\rho &= \{t:\ m_\rho(t)\in\mathcal{V}_\rho\},\\
\mathcal{O}_\rho &= \bigl\{[a,b]\in\mathcal{O}_\rho^{\mathrm{cand}}:\ |m_\rho(b)-m_\rho(a)|\geq\Delta_\rho^{\min},\ b-a+1\geq L_{\min}\bigr\}.
\end{aligned}
\label{eq:state_change}
\end{equation}
Separating reusable templates from their body-part-specific operators lets the registry extend beyond the conditions adapted from PoseScript and MotionScript. New pose and motion conditions can be expressed by composing existing operators and relations or by adding new entries to the registry.

\subsection{Answering 3D and 4D Queries}
\label{sec:method_inference}

\textbf{Query interpretation.} For both 3D and 4D queries, we provide $\Phi_{\mathrm{MLLM}}$ with a text-only prompt containing the query $q$, the category names and descriptions from the semantic memory $\mathcal{M}_{\text{semantic}}$, and a few in-context learning examples~\citep{brown2020language}, following skill-based prompting~\citep{tang2026wikiskill}. For a 4D query, the prompt also includes the registry's text cards. The model returns target categories $\mathcal{C}_q$ and, for a 4D query, a motion program $\Pi_q$ that defines operators by applying templates to body parts and links these operators with temporal relations.

\textbf{Spatial selection.} For both query types, the spatial mask $\mathbf{s}_q\in\{0,1\}^{N}$ selects canonical Gaussian $g_i$ if its stored category score $p_{ic}$ meets the threshold $\tau$ for any target category $c\in\mathcal{C}_q$:
\begin{equation}
s_{q,i}=\mathbf{1}\!\left[\exists\,c\in\mathcal{C}_q:\ p_{ic}\geq\tau\right].
\label{eq:spatial_selection}
\end{equation}

\textbf{Temporal selection.} For a 4D query, the executor runs the motion program $\Pi_q$ on the motion memory $\mathcal{M}_{\text{motion}}$. Each operator $\rho$ returns state frames $\mathcal{T}_\rho$ or change occurrences $\mathcal{O}_\rho$ from Eq.~\ref{eq:state_change}. The executor combines these outputs according to the temporal relations in $\Pi_q$ to obtain the matching interval set~$\mathcal{I}_q$. The registry's seven relations can require movements to co-occur, follow in order, or alternate, states to hold together, or a state to hold at the end of, throughout, or during a movement. For example, the program first identifies intervals when both upper arms move outward, and the \textit{StateAtEnd} relation then retains each interval whose last frame has both arms horizontal (relation definitions and execution example in Supp.~Sec.~\ref{app:relations}).

We convert the matching intervals in $\mathcal{I}_q$ into a temporal mask $\mathbf{h}_q\in\{0,1\}^{T}$:
\begin{equation}
h_{q,t}=\mathbf{1}\!\left[\exists\,I\in\mathcal{I}_q:\ t\in I\right].
\label{eq:temporal_selection}
\end{equation}
Combining the temporal mask $\mathbf{h}_q$ with the spatial mask $\mathbf{s}_q$ gives the spatiotemporal selection $\mathbf{Y}_q$:
\begin{equation}
\mathbf{Y}_q=[Y_{q,i,t}]_{i,t}\in\{0,1\}^{N\times T},\qquad Y_{q,i,t}=s_{q,i}\,h_{q,t}.
\label{eq:spatiotemporal_selection}
\end{equation}
Gaussian $g_i$ is selected at frame $t$ when both masks are active. Because $\mathbf{s}_q$ does not depend on $t$, $\mathbf{Y}_q$ selects the same region in every matching interval and none elsewhere. Selected categories explain which region is selected, while operator measurements and temporal relations explain when.

\section{Experiments}
\label{sec:experiments}
\suppressfloats[t]

\begin{table}[t]
\vspace{-\abovecaptionskip}
\input{tables/main/result_part_segmentation}
\vspace{-10pt}
\end{table}

\subsection{Evaluation Protocol}
\label{sec:benchmarks}
\textbf{Benchmarks.} To evaluate avatar referring segmentation, we introduce three benchmarks by augmenting existing human datasets with referring queries and human-annotated targets, following prior evaluations~\citep{he2025refersplat,li2025_4dlangsplat}. \RefThreeD{} extends ZJU-MoCap~\citep{peng2021neuralbody} with dense part annotations and 20 referring queries covering part names, appearance, wearing relations, and functions. \RefHuMManFourD{} extends HuMMan~\citep{cai2022humman} by converting its fine-grained body-part motion descriptions~\citep{zhang2023finemogen} into 20 referring queries, while \RefZJUFourD{} extends ZJU-MoCap with 20 queries in the same format. Construction and annotation details appear in Supp.~Secs.~\ref{app:bench_3d}--\ref{app:bench_4d}.

\textbf{Metrics.} For part segmentation, we report mean per-category IoU (mIoU), overall pixel accuracy (OA), and mean per-category recall (mAcc)~\citep{suzuki2025openvocab3dhuman}. For 3D referring segmentation, we report mIoU over queries~\citep{he2025refersplat}. For 4D referring segmentation, vIoU evaluates region accuracy and timing jointly, penalizing both incorrect masks and extra or missed active frames~\citep{li2025_4dlangsplat}. We additionally report tIoU, precision, and recall to assess temporal selection separately~\citep{fiebelman2025_4legs}. tIoU measures overlap between predicted and annotated active frames. Precision is the fraction of selected frames annotated as active, and recall the fraction of annotated active frames selected (Supp.~Sec.~\ref{app:metrics}).

\textbf{Implementation details.} Comparisons share avatar reconstructions. We use frozen Sapiens2-1B~\citep{khirodkar2026sapiens2} as $\Phi_{\mathrm{parse}}$ and Qwen3.8-27B~\citep{qwen2026qwen38} as $\Phi_{\mathrm{MLLM}}$. We set the score threshold to $\tau=0.4$ and the minimum interval length to $L_{\min}=4$ frames. Minimum-change thresholds $\Delta_\rho^{\min}$ are fixed per measurement type and shared across datasets. Their values, mask rendering, and comparison details appear in Supp.~Secs.~\ref{app:floors} and~\ref{app:protocols}.

\Needspace{8\baselineskip}
\subsection{Results on the 3D and 4D Evaluations}
\label{sec:results}
\begin{wraptable}[25]{r}{.50\linewidth}
\vspace{-\intextsep}
\vspace{-\abovecaptionskip}
\input{tables/main/result_referring_3d}
\par\vspace{-5pt}
\input{tables/main/result_joint_4d}
\vspace{-4pt}
\end{wraptable}
\textbf{Part segmentation.} Tab.~\ref{tab:parsing} compares our method with the human parsing methods 4D-DRESS~\citep{wang20244d} and OpenHuman3D/4D~\citep{suzuki2025openvocab3dhuman,suzuki2025openhuman4d} and the general-purpose 3D part segmentation model Find3D~\citep{ma2025find3d}. Our method achieves the highest mIoU on all nine avatars. Although designed for referring segmentation, our semantic memory thus also gives accurate part segmentation by assigning each Gaussian its highest-scoring category (Supp.~Fig.~\ref{fig:qual_parsing}).

\begin{figure}[!t]
\centering
\includegraphics[width=\linewidth]{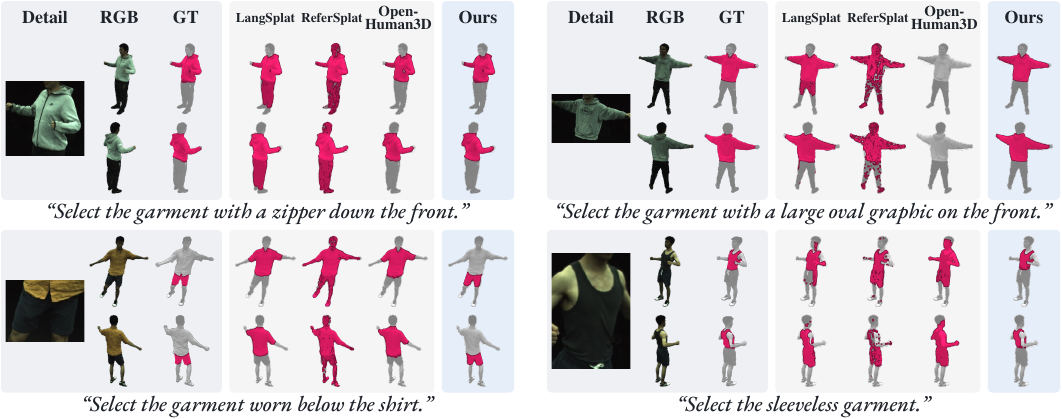}\par
\vspace{-5pt}
\caption{\textbf{Qualitative 3D referring segmentation on \RefThreeD.} Two held-out views and a detail crop per query; selected regions are \textcolor{wherered}{magenta}.}
\label{fig:qual3d}
\vspace{-10pt}
\end{figure}

\textbf{3D referring segmentation.} Tab.~\ref{tab:static3d} evaluates queries based on part names (Part), appearance (App.), and relations or functions (Rel.). \Ours{} achieves the highest mIoU across all three groups, with the largest margin on Rel. queries. Supplying the same Sapiens2 mask proposals to LangSplat~\citep{qin2024langsplat} and Dr.~Splat~\citep{kim2025drsplat} improves their Part and App. scores, but their CLIP-based matching still struggles with indirectly described targets. OpenHuman3D~\citep{suzuki2025openvocab3dhuman}, despite a CLIP encoder adapted to human parsing, and the general-purpose Find3D~\citep{ma2025find3d} also struggle with Rel. queries. With Sapiens2 supervision, ReferSplat~\citep{he2025refersplat}, which learns referring features from annotated expressions, performs best among the baselines on Rel. queries, yet still falls well short of our method. These results suggest that our MLLM-based selection from category names and descriptions resolves indirect references more accurately than feature matching, even when the same parser masks are provided. Fig.~\ref{fig:qual3d} further shows this difference for \emph{``the garment worn below the shirt''}, where LangSplat and OpenHuman3D select the reference shirt but our method selects the requested lower garment (more examples in Supp.~Sec.~\ref{app:qual_comparison}).

\begin{figure}[!t]
\centering
\includegraphics[width=\linewidth]{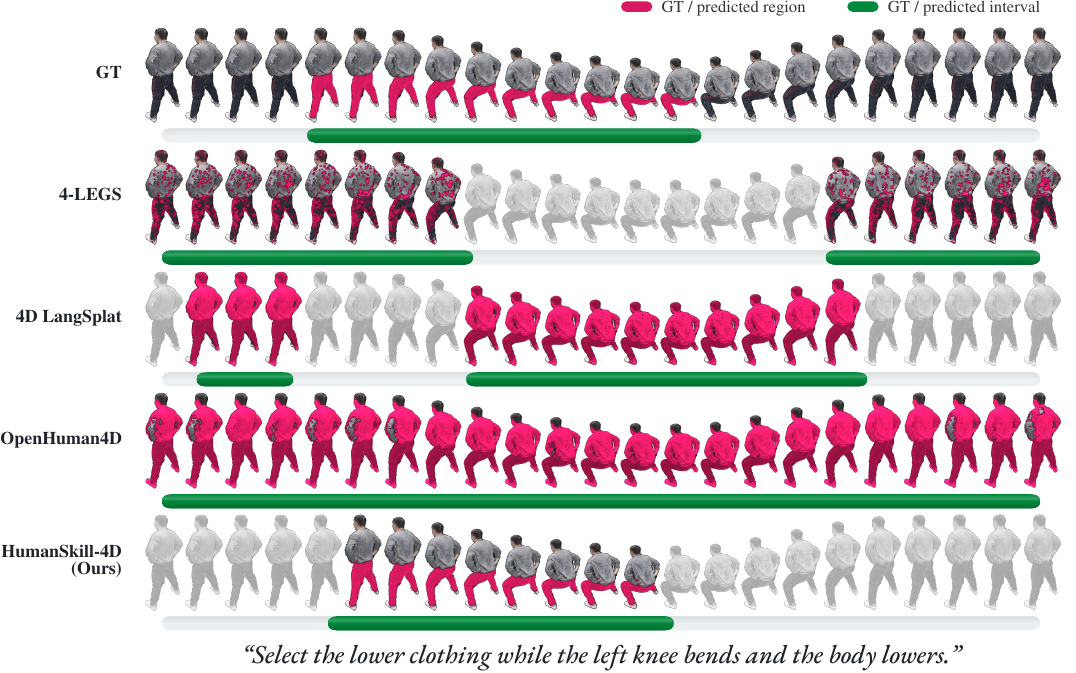}\par
\vspace{-8pt}
\caption{\textbf{Qualitative 4D referring segmentation on \RefHuMManFourD{}.} GT and predicted regions are \textcolor{wherered}{magenta} and intervals \textcolor{whenblue}{green}.}
\label{fig:qual4d}
\vspace{-2pt}
\makeatletter\def\@captype{table}\makeatother
\input{tables/main/result_cost}
\vspace{-12pt}
\end{figure}

\textbf{4D referring segmentation.} Tab.~\ref{tab:joint4d} reports joint region and interval selection results averaged over queries from \RefZJUFourD{} and \RefHuMManFourD{}. \Ours{} achieves the highest vIoU, tIoU, and temporal precision. \mbox{4-LEGS}~\citep{fiebelman2025_4legs} and 4D LangSplat~\citep{li2025_4dlangsplat}, which localize intervals from video features and MLLM-generated caption embeddings, respectively, have low temporal overlap and precision in this evaluation. OpenHuman4D~\citep{suzuki2025openhuman4d} instead keeps region predictions active throughout the clip, achieving 100\% temporal recall but lowering tIoU and joint vIoU by including inactive frames. Fig.~\ref{fig:qual4d} further shows spatial errors, where 4D LangSplat and OpenHuman4D select unrelated upper-body regions for a lower-clothing query (more examples in Supp.~Sec.~\ref{app:qual_temporal}). \Ours{} addresses both sources of error by resolving target categories through the semantic memory and executing MLLM-composed motion programs on the driving joint trajectories.

\subsection{Ablations and Computational Analysis}
\label{sec:ablations}

\textbf{Choice of MLLM.} To test how the choice of MLLM affects accuracy, we replace $\Phi_{\mathrm{MLLM}}$ in both description generation and query interpretation, with image inputs and queries fixed (Tab.~\ref{tab:ablations}, left). Qwen3.8 ties for the highest score in both 3D and 4D, and Qwen3.6 performs similarly, whereas Gemma 4 matches Qwen3.8's 4D vIoU but has much lower 3D mIoU. We use Qwen3.8 by default and report smaller MLLMs in Supp.~Sec.~\ref{app:ablations}.

\begin{table}[!t]
\centering
\caption{\textbf{Ablations} on MLLM choice (\textbf{left}), semantic category selector (\textbf{middle}), and score threshold $\tau$ (\textbf{right}). We report 3D mIoU and 4D vIoU (\%).}
\label{tab:ablations}
\vspace{-4pt}
\input{tables/main/ablation_llm_variants}
\hfill
\input{tables/main/ablation_selector}
\hfill
\input{tables/main/ablation_threshold}
\vspace{-9pt}
\end{table}

\textbf{Semantic category selector.} We examine whether text-embedding retrieval can replace MLLM-based category selection. CLIP~\citep{cherti2023scaling}, E5~\citep{wang2024e5mistral}, and JINA-v5~\citep{akram2026jinav5text} rank the same category descriptions and select the top three. We keep the semantic memory fixed and, for 4D queries, use the same temporal predictions. All three retrieval methods yield lower 3D mIoU and 4D vIoU than the MLLM (Tab.~\ref{tab:ablations}, middle). Qualitative examples (Supp.~Sec.~\ref{app:qual_comparison}) show JINA selecting the patched garment correctly but adding a forearm when asked to select both hands. These results support MLLM-based selection, because the selected set must contain the target categories and exclude those the query does not request, which a fixed top-three retrieval cannot ensure.

\textbf{Score threshold.} We test how sensitive segmentation accuracy is to $\tau$, keeping the memory, selected categories, and intervals fixed (Tab.~\ref{tab:ablations}, right). Performance remains stable for $\tau=0.2$--$0.5$, including our default $0.4$. At $\tau=0.1$, both mIoU and vIoU drop slightly, consistent with weakly supported Gaussians leaking across part or garment boundaries.

\textbf{Computational costs.} Tab.~\ref{tab:cost} compares the time to the first query and the storage retained for later queries (Supp.~Sec.~\ref{app:components}). LangSplat, ReferSplat, 4-LEGS, and 4D LangSplat optimize features for each avatar, whereas \Ours{} only registers the frozen parser's labels to the Gaussians, generates category descriptions with the frozen MLLM, and stores the joint trajectories. Our time to the first query is $\sim$4 min in 3D and $\sim$8 min in 4D, the shortest among the compared methods. The resulting memories occupy about 5 MB in 3D, comparable to LangSplat and ReferSplat, and 10 MB in 4D, nearly $10\times$ smaller than the most compact 4D baseline. \Ours{} thus achieves the highest accuracy with only minutes of setup and no per-avatar training.

\FloatBarrier
\raggedbottom
\section{Conclusion}
\label{sec:conclusion}
\looseness=-1
We introduced avatar referring segmentation, which selects avatar regions and the intervals satisfying queried pose or motion conditions, and presented \Ours{}, a training-free framework that gives the structure of Gaussian avatars to a frozen MLLM as a skill. To address region ambiguity, its semantic memory links body-part and garment categories and appearance descriptions to persistent Gaussian regions, enabling direct and indirect region references. To address motion ambiguity, the MLLM composes geometric operators adapted from PoseScript and MotionScript into programs executed on the driving joint trajectories. Because the MLLM chooses categories and operators rather than regions or intervals, its selections are traceable, and new motion conditions only require composing existing operators and relations or adding new ones. Experiments on our three human-annotated benchmarks show that \Ours{} selects regions and intervals most accurately, with compact memories and only minutes of setup per avatar. Limitations and future directions are discussed in Supp.~Sec.~\ref{app:limits}.

\flushbottom

\clearpage
\section*{AI use statement}
We used AI assistants under our guidance and supervision to debug code, search for relevant papers, check grammar, and improve readability and concision. We also used them to format LaTeX tables and adjust figure layouts and output formats. The MLLMs $\Phi_{\mathrm{MLLM}}$ were used only in our method and ablations.

\begingroup
\setlength{\bibsep}{6.5pt plus 4pt minus 2pt}
\bibliography{bibliography/refs}
\endgroup
\bibliographystyle{style/iclr2027_conference}

\clearpage
\appendix
\setcounter{figure}{0}
\setcounter{table}{0}
\renewcommand{\thefigure}{S\arabic{figure}}
\renewcommand{\thetable}{S\arabic{table}}
\renewcommand{\theHfigure}{appendix.\arabic{figure}}
\renewcommand{\theHtable}{appendix.\arabic{table}}
\counterwithin{equation}{section}
\renewcommand{\theHequation}{appendix.\thesection.\arabic{equation}}
\input{appendix}

\end{document}

%% file: tables/main/result_part_segmentation.tex
\caption{\textbf{Part segmentation on \RefThreeD.} OA, mAcc, and mIoU (\%) over 16 categories.}
\label{tab:parsing}
\centering
\small
\setlength{\tabcolsep}{3pt}
\begin{tabularx}{\linewidth}{@{}ll*{9}{>{\centering\arraybackslash}X}ccc@{}}
\toprule
\multicolumn{1}{c}{\multirow{2}{*}{Method}} & \multicolumn{1}{c}{\multirow{2}{*}{Venue}} & \multicolumn{9}{c}{mIoU per avatar} & \multicolumn{3}{c}{Overall} \\
\cmidrule(lr){3-11}\cmidrule(l){12-14}
& & 313 & 315 & 377 & 386 & 387 & 390 & 392 & 393 & 394 & OA & mAcc & mIoU \\
\midrule
4D-DRESS & CVPR'24 & 35.9 & 33.9 & 34.7 & 24.7 & 27.5 & 20.0 & 37.5 & 24.9 & 24.9 & 78.6 & 35.7 & 31.2 \\
OpenHuman3D & 3DV'25 & 47.7 & 44.8 & 32.3 & 35.6 & 30.9 & 25.9 & 44.3 & 45.9 & 26.9 & 66.6 & 51.8 & 41.2 \\
OpenHuman4D & BMVC'25 & 48.0 & 37.8 & 40.3 & 31.8 & 27.9 & 26.0 & 43.6 & 41.9 & 29.6 & 77.6 & 52.8 & 40.0 \\
Find3D & ICCV'25 & 17.0 & 13.3 & 13.1 & 13.4 & 8.8 & 8.1 & 13.5 & 12.1 & 10.7 & 14.9 & 31.5 & 12.7 \\
\midrule
\rowcolor{oursrow}\multicolumn{2}{@{}l}{\textbf{\Ours{}}} & \textbf{73.4} & \textbf{77.6} & \textbf{74.5} & \textbf{65.6} & \textbf{52.1} & \textbf{50.4} & \textbf{65.9} & \textbf{67.8} & \textbf{48.0} & \textbf{94.2} & \textbf{79.5} & \textbf{75.0} \\
\bottomrule
\end{tabularx}

%% file: tables/main/result_referring_3d.tex
\caption{\textbf{3D referring segmentation on \RefThreeD.} mIoU (\%) on part, appearance (App.), and relation/function (Rel.) queries; $^\dagger$~uses our parser $\Phi_{\mathrm{parse}}$ (i.e., Sapiens2).}
\label{tab:static3d}
\centering
\small
\setlength{\tabcolsep}{3pt}
\begin{tabularx}{\linewidth}{@{}l*{4}{>{\centering\arraybackslash}X}@{}}
\toprule
Method & Part & App. & Rel. & All \\
\midrule
LangSplat & 57.0 & 68.0 & 4.1 & 56.2 \\
LangSplat$^\dagger$ & 71.3 & 72.2 & 4.8 & 61.9 \\
ReferSplat & 28.6 & 47.6 & 5.1 & 37.4 \\
ReferSplat$^\dagger$ & 52.1 & 35.4 & 12.7 & 35.3 \\
OpenHuman3D & 40.3 & 54.9 & 3.1 & 44.2 \\
Dr.~Splat & 8.5 & 16.2 & 9.1 & 13.6 \\
Dr.~Splat$^\dagger$ & 54.5 & 44.0 & 2.5 & 39.9 \\
Find3D & 26.9 & 40.6 & 6.4 & 32.7 \\
\midrule
\rowcolor{oursrow}\textbf{\Ours{}} & \textbf{78.4} & \textbf{89.9} & \textbf{55.1} & \textbf{82.4} \\
\bottomrule
\end{tabularx}

%% file: tables/main/result_joint_4d.tex
\caption{\textbf{4D referring segmentation on \RefZJUFourD{} and \mbox{\RefHuMManFourD{}}} (\%), averaged over all queries.}
\label{tab:joint4d}
\centering
\small
\setlength{\tabcolsep}{3pt}
\begin{tabularx}{\linewidth}{@{}l*{4}{>{\centering\arraybackslash}X}@{}}
\toprule
Method & vIoU & tIoU & Prec & Rec\\
\midrule
4-LEGS & 1.9 & 13.6 & 17.6 & 35.7 \\
4D LangSplat & 4.9 & 17.7 & 22.6 & 45.3 \\
OpenHuman4D & 5.4 & 22.1 & 22.1 & \textbf{100.0} \\
\midrule
\rowcolor{oursrow}\textbf{\Ours{}} & \textbf{45.0} & \textbf{62.2} & \textbf{73.4} & 86.0 \\
\bottomrule
\end{tabularx}

%% file: tables/main/ablation_llm_variants.tex
\begin{minipage}[t]{.355\linewidth}
\vspace{0pt}
\centering
\small
\setlength{\tabcolsep}{3pt}
\begin{tabularx}{\linewidth}{@{}l*{2}{>{\centering\arraybackslash}X}@{}}
\toprule
MLLM & mIoU & vIoU \\
\midrule
Qwen3.6-27B & \textbf{82.4} & 44.3 \\
Gemma 4-31B & 55.3 & \textbf{45.0} \\
Mistral 3.2-24B & 58.6 & 38.8 \\
\rowcolor{oursrow}Qwen3.8-27B & \textbf{82.4} & \textbf{45.0} \\
\bottomrule
\end{tabularx}
\end{minipage}

%% file: tables/main/ablation_selector.tex
\begin{minipage}[t]{.35\linewidth}
\vspace{0pt}
\centering
\small
\setlength{\tabcolsep}{3pt}
\begin{tabularx}{\linewidth}{@{}l*{2}{>{\centering\arraybackslash}X}@{}}
\toprule
Selector & mIoU & vIoU \\
\midrule
CLIP & 34.9 & 13.1 \\
E5 & 41.4 & 13.3 \\
JINA-v5 & 63.1 & 19.4 \\
\rowcolor{oursrow}Qwen3.8-27B & \textbf{82.4} & \textbf{45.0} \\
\bottomrule
\end{tabularx}
\end{minipage}

%% file: tables/main/ablation_threshold.tex
\begin{minipage}[t]{.23\linewidth}
\vspace{0pt}
\centering
\small
\setlength{\tabcolsep}{3pt}
\begin{tabularx}{\linewidth}{@{}l*{2}{>{\centering\arraybackslash}X}@{}}
\toprule
$\tau$ & mIoU & vIoU \\
\midrule
0.1 & 80.7 & 43.2 \\
0.2 & 82.2 & 44.6 \\
0.3 & \textbf{82.4} & 44.9 \\
\rowcolor{oursrow}\textbf{0.4} & \textbf{82.4} & \textbf{45.0} \\
0.5 & 81.9 & 44.8 \\
\bottomrule
\end{tabularx}
\end{minipage}

%% file: tables/main/result_cost.tex
\caption{\textbf{Computational costs} for one new avatar in 3D (left) and 4D (right). Train.: per-avatar training; Time: minutes from a completed avatar to the first query's masks; Stor.: per-avatar data retained for querying (MB); Score: 3D mIoU or 4D vIoU over all benchmark queries (\%).}
\label{tab:cost}
\centering
\small
\setlength{\tabcolsep}{3pt}
\begin{minipage}[t]{.485\linewidth}
\begin{tabularx}{\linewidth}{@{}Xcccc@{}}
\toprule
Method & Train. & Time & Stor. & Score \\
\midrule
LangSplat & \checkmark & $\sim$75 & $\sim$2.4 & 56.2 \\
ReferSplat & \checkmark & $\sim$128 & $\sim$4.4 & 37.4 \\
OpenHuman3D & $\times$ & $\sim$62 & $\sim$260 & 44.2 \\
\rowcolor{oursrow}\textbf{\Ours{}} & $\times$ & $\sim$4 & $\sim$5 & \textbf{82.4} \\
\bottomrule
\end{tabularx}
\end{minipage}\hfill
\begin{minipage}[t]{.485\linewidth}
\begin{tabularx}{\linewidth}{@{}Xcccc@{}}
\toprule
Method & Train. & Time & Stor. & Score \\
\midrule
4-LEGS & \checkmark & $\sim$369 & $\sim$2{,}900 & 1.9 \\
4D LangSplat & \checkmark & $\sim$220 & $\sim$94 & 4.9 \\
OpenHuman4D & $\times$ & $\sim$32 & $\sim$1{,}900 & 5.4 \\
\rowcolor{oursrow}\textbf{\Ours{}} & $\times$ & $\sim$8 & $\sim$10 & \textbf{45.0} \\
\bottomrule
\end{tabularx}
\end{minipage}

%% file: appendix.tex
\section*{Supplementary material}
\label{app:overview}

\newcommand{\appentry}[2]{\medskip\noindent\hyperref[#1]{\textcolor{black}{\textbf{\ref*{#1}.\enspace #2}}}\enspace\dotfill\enspace\hyperref[#1]{\textcolor{black}{p.~\pageref*{#1}}}\par\smallskip}
\newcommand{\subentry}[2]{\noindent\hspace{1.5em}\hyperref[#1]{\textcolor{black}{\ref*{#1}\enspace #2}}\enspace\dotfill\enspace\hyperref[#1]{\textcolor{black}{p.~\pageref*{#1}}}\par\smallskip}

\appentry{app:method}{Method details}
\subentry{app:categories}{Semantic categories}
\subentry{app:registry}{Operator templates}
\subentry{app:floors}{State frames and change occurrences}
\subentry{app:relations}{Combining operators with temporal relations}
\subentry{app:llm_io}{MLLM inputs and outputs}

\appentry{app:evaluation}{Benchmarks}
\subentry{app:bench_3d}{\RefThreeD}
\subentry{app:bench_4d}{4D referring evaluation}

\appentry{app:comparison}{Evaluation details}
\subentry{app:metrics}{Metrics}
\subentry{app:protocols}{Implementation and comparison protocols}
\subentry{app:components}{Computational cost accounting}

\appentry{app:results}{Additional results}
\subentry{app:ablations}{Smaller MLLMs}
\subentry{app:qual_parsing}{Qualitative examples}

\appentry{app:limits}{Limitations and future directions}

\appentry{app:query_list}{Complete query list}
\appentry{app:prompts}{MLLM instructions}
\clearpage

\section{Method details}
\label{app:method}

This section expands the definitions and execution rules introduced in Sec.~\ref{sec:method} of the main paper.

\textbf{Running query.} We use the query \emph{``Select the upper clothing while both upper arms lift outward until they are parallel to the ground''} from Fig.~\ref{fig:queries}(b) of the main paper as a running example to show how motion operators and temporal relations determine the selected intervals.

\subsection{Semantic categories}
\label{app:categories}
The semantic memory uses a fixed set $\mathcal{K}$ of 16 categories, obtained by mapping the labels of $\Phi_{\mathrm{parse}}$ (Sapiens2~\citep{khirodkar2026sapiens2}): \texttt{hair}, \texttt{face\_neck}, \texttt{torso}, \texttt{left\_arm}, \texttt{right\_arm}, \texttt{left\_hand}, \texttt{right\_hand}, \texttt{left\_leg}, \texttt{right\_leg}, \texttt{upper\_clothing}, \texttt{lower\_clothing}, \texttt{left\_shoe}, \texttt{right\_shoe}, \texttt{left\_sock}, \texttt{right\_sock}, and \texttt{eyeglass}. The mapping merges the upper and lower arm into \texttt{left\_arm} or \texttt{right\_arm}, the upper leg, lower leg, and foot into \texttt{left\_leg} or \texttt{right\_leg}, and the lips into \texttt{face\_neck}.

\subsection{Operator templates}
\label{app:registry}

Motion operators are built from the templates in the operator registry $\mathcal{R}$. A template specifies what to measure and what condition to check. For example, the \emph{near} and \emph{approach} templates both measure the distance between two body points. The \emph{near} template checks whether the points are close at a frame (a state), while \emph{approach} checks whether their distance decreases over time (a change).

\textbf{Operator registry.} Tab.~\ref{tab:registry} lists the templates of the registry $\mathcal{R}$ by measurement type, with state and change templates in separate columns. We define these templates by adapting the geometric pose and motion codes of PoseScript~\citep{delmas2022posescript} and MotionScript~\citep{yazdian2025motionscript} to conditions over SMPL trajectories. For example, forward and backward head tilt use the \emph{move forward} and \emph{move backward} templates applied to the neck--head segment. New pose and motion conditions can be supported by composing existing operators and relations or by adding new registry entries, each defined by its rule and text card.

\input{tables/appendix/reference_registry}

\textbf{Applying templates.} To apply one of these templates, we specify a compatible body part, which determines the SMPL joints used for the measurement. For an elbow, the \emph{bend} template measures the shoulder--elbow--wrist angle, and for an upper arm, the \emph{horizontal} template measures the angle of the shoulder--elbow segment from vertical. Other templates use a point and a reference, as the \emph{above} template does for ``the hand is above the head''.

\input{figures/appendix/running_templates}
\textbf{Running example.} The phrase \emph{lift outward} selects the \emph{move outward} template (Fig.~\ref{fig:f01_templates}, left) of the relative-position type in Tab.~\ref{tab:registry}, which tracks the elbow's sideways offset from the shoulder, toward that arm's own side, along the left--right axis of the fixed frame. The phrase \emph{parallel to the ground} selects the \emph{horizontal} template (Fig.~\ref{fig:f01_templates}, right), which checks whether the shoulder--elbow segment is horizontal. Applying both templates to each upper arm gives four operators: the left and right outward movements $\rho_1$ and $\rho_2$, and the left and right horizontal states $\rho_3$ and $\rho_4$.

\par\vspace{1.5\baselineskip}\WFclear
\subsection{State frames and change occurrences}
\label{app:floors}

\textbf{State ranges.} A state operator checks at every frame whether its measurement lies in its valid range $\mathcal{V}_\rho$, and the ranges follow the posecodes of PoseScript~\citep{delmas2022posescript} and MotionScript~\citep{yazdian2025motionscript}. For joint angles, where $180^\circ$ is straight, we use their bin boundaries at 45$^\circ$, 75$^\circ$, 105$^\circ$, 135$^\circ$, and 160$^\circ$ as fixed values rather than PoseScript's randomized binning. The \emph{straight} template covers angles above 160$^\circ$, the five bent degrees in Tab.~\ref{tab:registry} take the other five bins from \emph{slight} (135$^\circ$--160$^\circ$) to \emph{full} (45$^\circ$ or less), and the \emph{broad bent} template covers all five. A point is above, below, in front of, behind, left of, or right of its reference only beyond a 0.15\,m dead zone along the corresponding axis of the fixed frame. The root pitch is neutral within 5$^\circ$ of upright. A segment is vertical within 10$^\circ$ of vertical and horizontal beyond 80$^\circ$ from vertical. Two points are \emph{near} within 0.2\,m and \emph{far} beyond 0.8\,m, the closest and widest of the posecode distance bins. A foot or ankle is near the lowest body joint within 0.1\,m.

\textbf{Candidate intervals.} A change operator uses the frame-to-frame difference $m_\rho(t)-m_\rho(t-1)$ to split its measurement, such as the elbow angle for the \emph{bend} template, into runs in which it increases or decreases, and keeps the runs in the direction of its template as candidates. A candidate starts at the frame before its first change and ends at the frame before the first change in the opposite direction, where the next run starts; frames without change in between stay in the candidate. If no opposite change follows, the candidate ends at its last change, so a plateau at the end of the sequence is excluded. For example, if the elbow angle drops at every frame from 10 to 25, holds for three frames, drops again through frame 40, and rises at frame 41, the \emph{bend} template yields the single candidate $[9,40]$. A candidate becomes an occurrence when it passes the two thresholds of Eq.~\ref{eq:state_change} of the main paper, a minimum total change $\Delta_\rho^{\min}$ and a minimum length of $L_{\min}$ frames. Because the candidate fixes the direction, the absolute value in Eq.~\ref{eq:state_change} measures only the amount of change.

\textbf{Minimum-change thresholds.} For each measurement type with change templates, we set $\Delta_\rho^{\min}$ to the 99.5th percentile of the total change over all runs of at most 3 frames in HuMMan sequences not used in our evaluation. The resulting values, shared by HuMMan and ZJU-MoCap, are 1.49$^\circ$ for joint angles, 8.6\,mm for relative position, 13.8\,mm for displacement, 4.94$^\circ$ for root rotation, 6.16$^\circ$ for segment tilt, and 15.1\,mm for distance. We set $L_{\min}=4$ frames on both datasets to exclude runs of at most 3 frames, even when their total change exceeds $\Delta_\rho^{\min}$.

\textbf{Change magnitude.} Following the angular motion codes of MotionScript~\citep{yazdian2025motionscript}, we assign each occurrence a magnitude from the number of bins between its start and end values. One or two bins make the change \emph{slight}, three \emph{moderate}, and four or more \emph{significant}. Occurrences that change by less than PoseScript's 5$^\circ$ binning noise for angles and orientations receive no magnitude. Only joint-angle and root-rotation changes receive a magnitude, and root rotation counts the bins of its own rotation axis. Magnitudes thus describe changes rather than states, so \emph{slightly bent} is the \emph{slight} bent degree in Tab.~\ref{tab:registry}, whereas \emph{slightly bending} is a bend of \emph{slight} magnitude.

\textbf{Reference frames and body shape.} All measurements use joints of the default SMPL shape~\citep{loper2015smpl} rather than subject-specific shape coefficients, which keeps the thresholds comparable across subjects. The fixed frame is set at the first frame (Sec.~\ref{sec:method_operators} of the main paper) and does not rotate when the body later turns. Root-local measurements follow the body root, and gravity-frame measurements use only the vertical axis (Tab.~\ref{tab:registry}).

\input{figures/appendix/running_measurements}

\textbf{Running example.} The outward-motion operators $\rho_1$ and $\rho_2$ measure this offset for the left and right elbows (Fig.~\ref{fig:f01_measure}, top). Only two outward candidates per arm change by at least 8.6\,mm, one in the middle of the clip and one at its end. The state operators $\rho_3$ and $\rho_4$ return the frames in which each shoulder--elbow segment is more than 80$^\circ$ from vertical (Fig.~\ref{fig:f01_measure}, bottom), which holds at the start of the clip and again as the first lift ends.

\FloatBarrier
\subsection{Combining operators with temporal relations}
\label{app:relations}

State operators return the set of frames in which a state holds, and change operators return a list of occurrences, each an interval covering one movement. Tab.~\ref{tab:relations} lists the registry's temporal relations that combine them in a motion program, with the input types each relation accepts and the output it returns.

\input{tables/appendix/reference_relations}
\input{figures/appendix/f01_trace}
\FloatBarrier

\textbf{Running example.} For this query, $\Phi_{\mathrm{MLLM}}$ returns the motion program
\begin{equation*}
\Pi_q=\text{StateAtEnd}\bigl(\text{CoOccurrence}(\rho_1,\rho_2),\ \text{AllStates}(\rho_3,\rho_4)\bigr),
\end{equation*}
where $\rho_1$ and $\rho_2$ are the outward movements of the left and right upper arms and $\rho_3$ and $\rho_4$ are their horizontal states (Sec.~\ref{app:registry}). In Fig.~\ref{fig:f01_trace}, \textit{CoOccurrence} keeps the two co-occurrences in which both upper arms move outward, frames 37--56 and 95--102, and \textit{AllStates} keeps the frames in which both are horizontal. \textit{StateAtEnd} keeps only the first co-occurrence, whose last frame 56 lies in a shared horizontal state; the second ends at frame 102, where neither arm is horizontal. The program therefore returns frames 37--56, the full interval of the lift that reaches horizontal. Because each of these records can be inspected, an error can be traced to the step that produced it.

\FloatBarrier
\subsection{MLLM inputs and outputs}
\label{app:llm_io}

In the query interpretation of Sec.~\ref{sec:method_inference} of the main paper, $\Phi_{\mathrm{MLLM}}$ acts as a 3D selector or a 4D compiler. The selector receives three in-context learning (ICL) examples~\citep{brown2020language} and the compiler five, and Sec.~\ref{app:prompts} gives the core of each instruction with an ICL example.

\textbf{3D selector.} For a 3D query, the selector receives the query and the names and descriptions of $\mathcal{K}$. It returns the target categories $\mathcal{C}_q\subseteq\mathcal{K}$ or an empty list.

\textbf{4D compiler.} For a 4D query, the compiler receives the query, the same category names and descriptions, the text cards of the templates in the registry $\mathcal{R}$ and of the relations in Tab.~\ref{tab:relations}. A template card tells $\Phi_{\mathrm{MLLM}}$ what the template measures, whether it checks a state or a change, which body parts and coordinate frame it uses, and which query descriptions it fits. A relation card states the relation's meaning, the inputs it accepts, and the output it returns. The compiler returns the target categories $\mathcal{C}_q$ and a motion program $\Pi_q$, or declares the query unsupported. For the running query, the \emph{move outward} and \emph{horizontal} cards match \emph{lift outward} and \emph{parallel to the ground} on the upper arms, and the \textit{CoOccurrence}, \textit{AllStates}, and \textit{StateAtEnd} cards let the compiler require both arms to move together and to end horizontal (Tab.~\ref{tab:llm_cards}), which yields the program in Sec.~\ref{app:relations}.

\input{tables/appendix/reference_llm_cards}

\FloatBarrier
\section{Benchmarks}
\label{app:evaluation}

\subsection{\RefThreeD}
\label{app:bench_3d}

\RefThreeD{} extends ZJU-MoCap~\citep{peng2021neuralbody} with two forms of human annotation. For part segmentation, nine subjects are annotated at eight frames in four held-out cameras, yielding 288 ground-truth (GT) masks over the 16 categories of Sec.~\ref{app:categories}. For 3D referring segmentation, each query is evaluated on the 32 held-out images of its subject (Sec.~\ref{app:query_list}).

\subsection{4D referring evaluation}
\label{app:bench_4d}

\RefHuMManFourD{} and \RefZJUFourD{} each contain 20 queries (Sec.~\ref{app:query_list}). In both benchmarks, each query has a target mask in two held-out cameras for every active frame. For HuMMan, we select sequences showing movements of specific body parts with clearly identifiable starts and ends, and annotators convert the provided fine-grained motion descriptions~\citep{zhang2023finemogen} into referring queries. Annotators then review each clip and manually determine the intervals satisfying the specified pose or motion conditions. ZJU-MoCap queries use the same format, and annotators determine their intervals in the same way.

\section{Evaluation details}
\label{app:comparison}

\subsection{Metrics}
\label{app:metrics}

\textbf{mIoU in part segmentation.}\label{app:parsing_metric} For category $c$, we accumulate the prediction--GT intersection and union pixel counts, $n_c^{\cap}$ and $n_c^{\cup}$, over the evaluated images~\citep{long2015fcn}. The score averages the IoU over the categories $\mathcal{C}_{+}$ that appear in the GT or the prediction:
\begin{equation}
\mathcal{C}_{+}=\{c:n_c^{\cup}>0\},\qquad
\mathrm{mIoU}=\frac{1}{|\mathcal{C}_{+}|}\sum_{c\in\mathcal{C}_{+}}\frac{n_c^{\cap}}{n_c^{\cup}}.
\label{eq:parsing_miou}
\end{equation}

\textbf{OA and mAcc in part segmentation.} OA is the fraction of annotated pixels assigned the correct category, and mAcc is the mean per-category recall.

\textbf{IoU for a target mask.}\label{app:spatial_metric} For the predicted and GT Boolean masks $M^{\mathrm{pred}}$ and $M^{\mathrm{GT}}$ of an image, we use the region overlap $\mathcal{J}$ defined by DAVIS~\citep{perazzi2016davis}, with the empty-mask convention of its official evaluator, which scores an image 1 when both masks are empty:
\begin{equation}
\mathcal{J}(M^{\mathrm{pred}},M^{\mathrm{GT}})=
\begin{cases}
\dfrac{|M^{\mathrm{pred}}\cap M^{\mathrm{GT}}|}{|M^{\mathrm{pred}}\cup M^{\mathrm{GT}}|}, & |M^{\mathrm{pred}}\cup M^{\mathrm{GT}}|>0,\\[3pt]
1, & |M^{\mathrm{pred}}\cup M^{\mathrm{GT}}|=0.
\end{cases}
\label{eq:mask_iou}
\end{equation}

\textbf{mIoU in 3D referring segmentation.} We evaluate 3D referring segmentation over pixels with human part annotations. Each query $q$ in the 3D query set $\mathcal{Q}_{\mathrm{3D}}$ has held-out images $\mathcal{E}_q$, and each image $e$ has the predicted and GT masks $M^{\mathrm{pred}}_{qe}$ and $M^{\mathrm{GT}}_{qe}$. The score averages their overlap $\mathcal{J}$ (Eq.~\ref{eq:mask_iou}) over the held-out images of each query and then over the queries:
\begin{equation}
\mathrm{mIoU}=\frac{1}{|\mathcal{Q}_{\mathrm{3D}}|}\sum_{q\in\mathcal{Q}_{\mathrm{3D}}}
\frac{1}{|\mathcal{E}_q|}\sum_{e\in\mathcal{E}_q}\mathcal{J}(M^{\mathrm{pred}}_{qe},M^{\mathrm{GT}}_{qe}).
\label{eq:query_miou}
\end{equation}

\textbf{tIoU, precision, and recall.}\label{app:temporal_metric}\label{app:metric_4d} For query $q$, the clip frames $\mathcal{F}_q$ contain the predicted active frames $P_q$ and the GT-active frames $G_q$. We use the temporal measures reported by 4-LEGS~\citep{fiebelman2025_4legs}:
\begin{equation}
\mathrm{tIoU}_q = \frac{|P_q\cap G_q|}{|P_q\cup G_q|},\qquad
\mathrm{Prec}_q = \frac{|P_q\cap G_q|}{|P_q|},\qquad
\mathrm{Rec}_q = \frac{|P_q\cap G_q|}{|G_q|}.
\label{eq:frame_metrics}
\end{equation}
Precision decreases when the prediction includes frames outside the GT intervals, while recall decreases when it omits GT-active frames. We set a metric to 0 when its denominator is 0.

\textbf{vIoU in 4D referring segmentation.}\label{app:joint_metric} Following 4D LangSplat~\citep{li2025_4dlangsplat}, we score each frame by the mask IoU $\mathcal{J}$ rather than by the bounding-box IoU of 4-LEGS~\citep{fiebelman2025_4legs}. With the predicted and GT masks $M^{\mathrm{pred}}_{qvt}$ and $M^{\mathrm{GT}}_{qvt}$ for query $q$, held-out camera $v$, and frame $t$, the score for camera $v$ is
\begin{equation}
\mathrm{vIoU}_{qv}=\frac{\displaystyle\sum_{t\in P_q\cap G_q}
\mathcal{J}(M^{\mathrm{pred}}_{qvt},M^{\mathrm{GT}}_{qvt})}{|P_q\cup G_q|}.
\label{eq:mask_viou}
\end{equation}
Each frame in both sets contributes its own mask IoU $\mathcal{J}$, whereas extra frames ($P_q\setminus G_q$) and missed frames ($G_q\setminus P_q$) contribute 0. Because $\mathcal{J}\le 1$, $\mathrm{vIoU}_{qv}$ never exceeds $\mathrm{tIoU}_q$ and reaches it only when every overlapping frame matches the target mask exactly. For the 4D query set $\mathcal{Q}_{\mathrm{4D}}$, where query $q$ has held-out cameras $\mathcal{W}_q$, vIoU averages $\mathrm{vIoU}_{qv}$ over cameras and then over queries:
\begin{equation}
\mathrm{vIoU}=\frac{1}{|\mathcal{Q}_{\mathrm{4D}}|}\sum_{q\in\mathcal{Q}_{\mathrm{4D}}}
\frac{1}{|\mathcal{W}_q|}\sum_{v\in\mathcal{W}_q}\mathrm{vIoU}_{qv}.
\label{eq:mean_mask_viou}
\end{equation}

\subsection{Implementation and comparison protocols}
\label{app:protocols}

\textbf{Our configuration.} We use the frozen public releases of Sapiens2-1B\footnote{\href{https://huggingface.co/facebook/sapiens2-seg-1b}{\texttt{facebook/sapiens2-seg-1b}}}~\citep{khirodkar2026sapiens2} for $\Phi_{\mathrm{parse}}$ and Qwen3.8-27B\footnote{\href{https://huggingface.co/Qwen/Qwen3.8-27B}{\texttt{Qwen/Qwen3.8-27B}}}~\citep{qwen2026qwen38} for $\Phi_{\mathrm{MLLM}}$. Each avatar is reconstructed with 3DGS-Avatar~\citep{qian2024gsavatar} from eight training cameras at $512\times512$ resolution, using the SMPL~\citep{loper2015smpl} provided by each dataset. For the registration set $\mathcal{D}_{\text{reg}}$, we sample four cameras every 32 frames on ZJU-MoCap and six cameras every 2 frames on HuMMan.

\textbf{Masks from Gaussian selections.} For part segmentation, each Gaussian takes the category that received its largest accumulated rendering contribution, and each pixel takes the category whose Gaussians contribute most to it. For referring segmentation, the Gaussians selected by Eq.~\ref{eq:spatial_selection} of the main paper render as 1 and all others as 0. We map the rendering back to the original camera image and binarize the rendered values at $>0.7$, following OpenGaussian's cutoff~\citep{wu2024opengaussian}.

\textbf{Ablation models.} The MLLM ablation in Tab.~\ref{tab:ablations} (left) of the main paper uses Qwen3.8-27B, Qwen3.6-27B\footnote{\href{https://huggingface.co/Qwen/Qwen3.6-27B}{\texttt{Qwen/Qwen3.6-27B}}}~\citep{qwen2026qwen36}, Gemma 4-31B\footnote{\href{https://huggingface.co/google/gemma-4-31B-it}{\texttt{google/gemma-4-31B-it}}}~\citep{gemma2026gemma4}, and Mistral 3.2-24B\footnote{\href{https://huggingface.co/mistralai/Mistral-Small-3.2-24B-Instruct-2506}{\texttt{mistralai/Mistral-Small-3.2-24B-Instruct-2506}}}. The smaller MLLMs in Sec.~\ref{app:ablations} are Gemma 4 E2B\footnote{\href{https://huggingface.co/google/gemma-4-E2B-it}{\texttt{google/gemma-4-E2B-it}}}~\citep{gemma2026gemma4}, Ministral 3 8B\footnote{\href{https://huggingface.co/mistralai/Ministral-3-8B-Instruct-2512}{\texttt{mistralai/Ministral-3-8B-Instruct-2512}}}~\citep{liu2026ministral3}, Qwen3-VL-8B\footnote{\href{https://huggingface.co/Qwen/Qwen3-VL-8B-Instruct}{\texttt{Qwen/Qwen3-VL-8B-Instruct}}}~\citep{qwen2025qwen3vl}, and Phi-4\footnote{\href{https://huggingface.co/microsoft/Phi-4-multimodal-instruct}{\texttt{microsoft/Phi-4-multimodal-instruct}}}~\citep{microsoft2025phi4mini}. The selector ablation uses CLIP\footnote{\href{https://huggingface.co/laion/CLIP-ViT-B-16-laion2B-s34B-b88K}{\texttt{laion/CLIP-ViT-B-16-laion2B-s34B-b88K}}}~\citep{cherti2023scaling}, E5\footnote{\href{https://huggingface.co/intfloat/e5-mistral-7b-instruct}{\texttt{intfloat/e5-mistral-7b-instruct}}}~\citep{wang2024e5mistral}, and JINA-v5\footnote{\href{https://huggingface.co/jinaai/jina-embeddings-v5-text-small}{\texttt{jinaai/jina-embeddings-v5-text-small}}}~\citep{akram2026jinav5text}.

\textbf{Part segmentation.}\label{app:parsing_protocol} In Tab.~\ref{tab:parsing} of the main paper, all methods use the same reconstructed avatars and training observations. All methods are scored on the same 16 categories, and text-based methods receive all 16 names at once. Find3D~\citep{ma2025find3d} predicts point features aligned with SigLIP~\citep{zhai2023siglip} text features from a point cloud with coordinates, RGB, and normals. Following \citet{li2026chorus}, we give the released Find3D a point cloud of the canonical Gaussians, with their centers, their RGB averaged over training observations, and normals estimated from neighboring centers, and map its point labels back to the Gaussians. 4D-DRESS~\citep{wang20244d} parses mesh vertices, so we run it on a mesh fused from the avatar's rendered RGB and depth at each evaluated frame and give each pixel the label of the closest vertex of the mesh triangle it sees, mapped to our 16 categories.

\textbf{OpenHuman3D and OpenHuman4D.} OpenHuman3D~\citep{suzuki2025openvocab3dhuman} projects multi-view SAM~\citep{kirillov2023sam} region proposals onto the shared avatar and fuses their HumanCLIP features, which come from an Alpha-CLIP~\citep{sun2024alphaclip} image encoder fine-tuned for human parsing. OpenHuman4D~\citep{suzuki2025openhuman4d} additionally tracks these regions over time with SAM~2~\citep{ravi2025sam2}. Both label each region with the given text it matches best. For referring segmentation, we give them the full query and the word ``other'' and select the regions that match the query better than ``other''.

\textbf{3D referring segmentation.}\label{app:readout} All methods in Tab.~\ref{tab:static3d} of the main paper use the same reconstructed Gaussian avatars, queries, and human annotations. LangSplat~\citep{qin2024langsplat} encodes its SAM~\citep{kirillov2023sam} partitions with CLIP~\citep{radford2021clip}, compresses these features with a per-avatar autoencoder, and fits language fields to them. Dr.~Splat~\citep{kim2025drsplat} registers CLIP features on the shared Gaussians with its default settings and compresses them with product quantization; features gathered on posed Gaussians are stored on the corresponding canonical Gaussians. Find3D uses its official release as is. ReferSplat~\citep{he2025refersplat} learns referring features for each scene from phrases encoded with BERT~\citep{devlin2019bert} and paired with target masks, which we generate with Grounded-SAM~\citep{ren2024groundedsam} from each phrase and its training image. In the Sapiens2 variants, LangSplat and Dr.~Splat replace their own masks with the same Sapiens2 mask proposals, without class names, and ReferSplat replaces the Grounded-SAM masks with Sapiens2 category masks.

\textbf{4D referring segmentation.}\label{app:temporal} In Tab.~\ref{tab:joint4d} of the main paper, 4-LEGS~\citep{fiebelman2025_4legs} compresses ViCLIP~\citep{wang2024internvid} video-text features with an autoencoder, attaches them to the Gaussians, and scores their relevance to the query text over time. It returns the high-relevance frames as the query intervals and renders the high-relevance Gaussians into a mask with the same rendering and 0.7 cutoff used for all Gaussian selections. 4D LangSplat~\citep{li2025_4dlangsplat} has Qwen2-VL~\citep{wang2024qwen2vl} caption how each region changes over time. It compresses CLIP features of the regions and E5~\citep{wang2024e5mistral} embeddings of these captions with autoencoders, fits spatial and temporal language fields to them, and matches the query against both.

\input{tables/appendix/reference_method_components}

\subsection{Computational cost accounting}
\label{app:components}

\textbf{Training.} Train.\ marks per-avatar field or autoencoder training. OpenHuman3D/4D instead use HumanCLIP, fine-tuned once by its authors for human parsing and excluded from Time.

\textbf{Time.} Time starts with completed Gaussians, RGB views, cameras, and poses, before method-specific preparation, and ends when the first query's final masks are saved. It includes per-avatar segmentation, description generation, feature extraction, training, semantic registration, pose preparation, and query execution. Model loading and all inference calls are included for every method. All methods run on an NVIDIA A100 80GB.

\textbf{Storage.} Storage counts the per-avatar data retained for querying, listed under Stored information in Tab.~\ref{tab:components}. \emph{Features} are the language features each method stores for Gaussians or regions. \emph{Networks} are layers trained for each avatar, such as the autoencoders that decode compressed features, ReferSplat's layers that match Gaussian features to the query text, and 4-LEGS' layers that combine each Gaussian's feature with those of its neighbors at each frame. \emph{Masks} are OpenHuman's region masks, and \emph{scores}, \emph{text}, and \emph{motion} are our category scores, category descriptions, and joint trajectories. Storage excludes shared Gaussians, camera and pose inputs, pretrained model weights shared across avatars, training intermediates, and query outputs.

\section{Additional results}
\label{app:results}

\subsection{Smaller MLLMs}
\label{app:ablations}

\begingroup
\setlength{\columnsep}{5pt}
\begin{wraptable}[9]{r}{.295\linewidth}
\vspace{-\intextsep}
\vspace{-\abovecaptionskip}
\input{tables/appendix/ablation_llm_small}
\end{wraptable}
In Tab.~\ref{tab:llm_small}, we replace $\Phi_{\mathrm{MLLM}}$ with four smaller public MLLMs under the same protocol. In 3D, Qwen3-VL-8B comes close to Qwen3.8-27B and even exceeds the larger Gemma 4-31B and Mistral 3.2-24B in Tab.~\ref{tab:ablations} (left) of the main paper. In 4D, however, all four smaller models fall far below the larger MLLMs, suggesting that composing motion programs from motion descriptions demands more capacity than selecting categories.
\par
\endgroup

\FloatBarrier

\subsection{Qualitative examples}
\label{app:qual_parsing}

\begin{figure}[!ht]
\centering
\includegraphics[width=\linewidth]{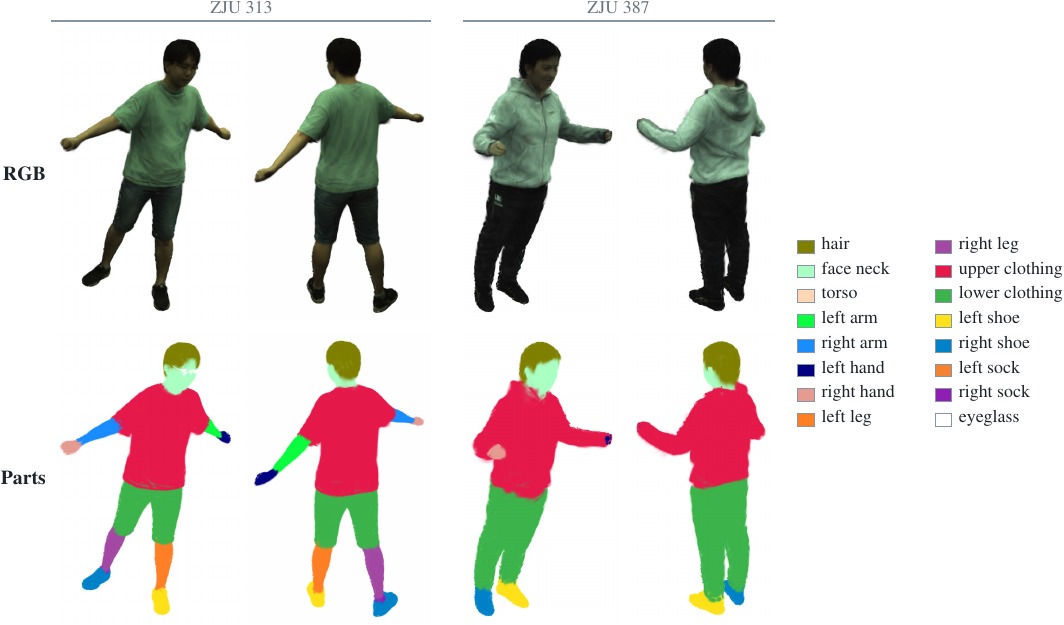}\par
\vspace{-7.5pt}
\caption{\textbf{Qualitative part segmentation on \RefThreeD{}.}}
\label{fig:qual_parsing}
\end{figure}

\textbf{Part segmentation.} Fig.~\ref{fig:qual_parsing} renders the stored category assignments of two avatars from held-out cameras. The labels stay on the same body parts as the view changes, because each label belongs to a canonical Gaussian rather than to a single image.

\begin{figure}[!ht]
\centering
\includegraphics[width=\linewidth]{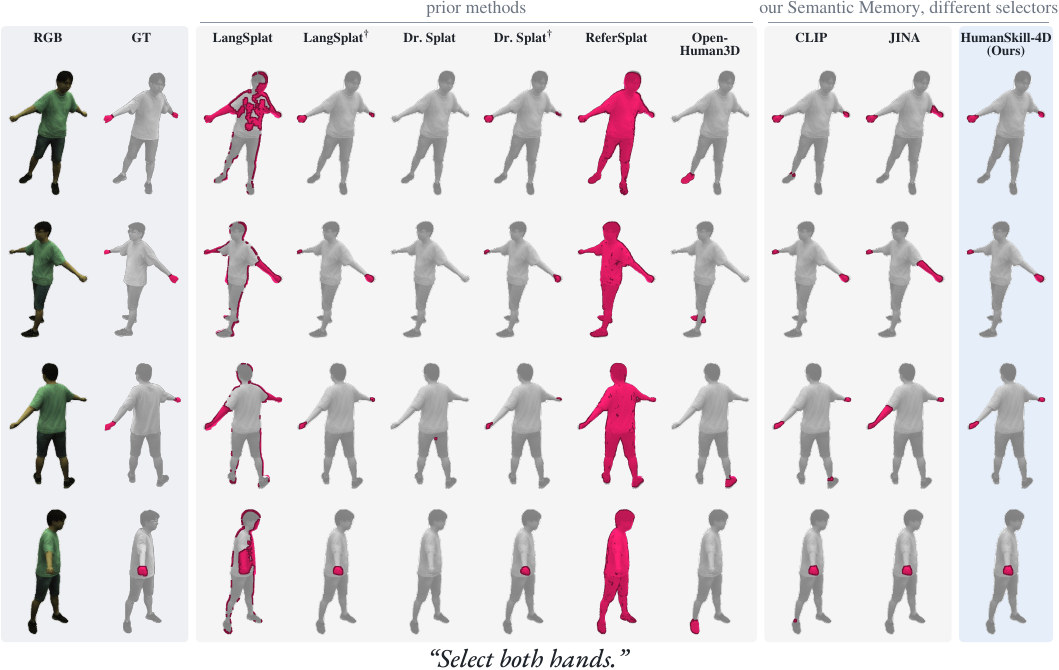}\par
\vspace{1.5pt}
\caption{\textbf{Qualitative 3D referring segmentation on \RefThreeD{}.} $^\dagger$~uses Sapiens2 masks.}
\label{fig:qual_comparison}
\end{figure}

\begin{figure}[!ht]
\centering
\includegraphics[width=\linewidth]{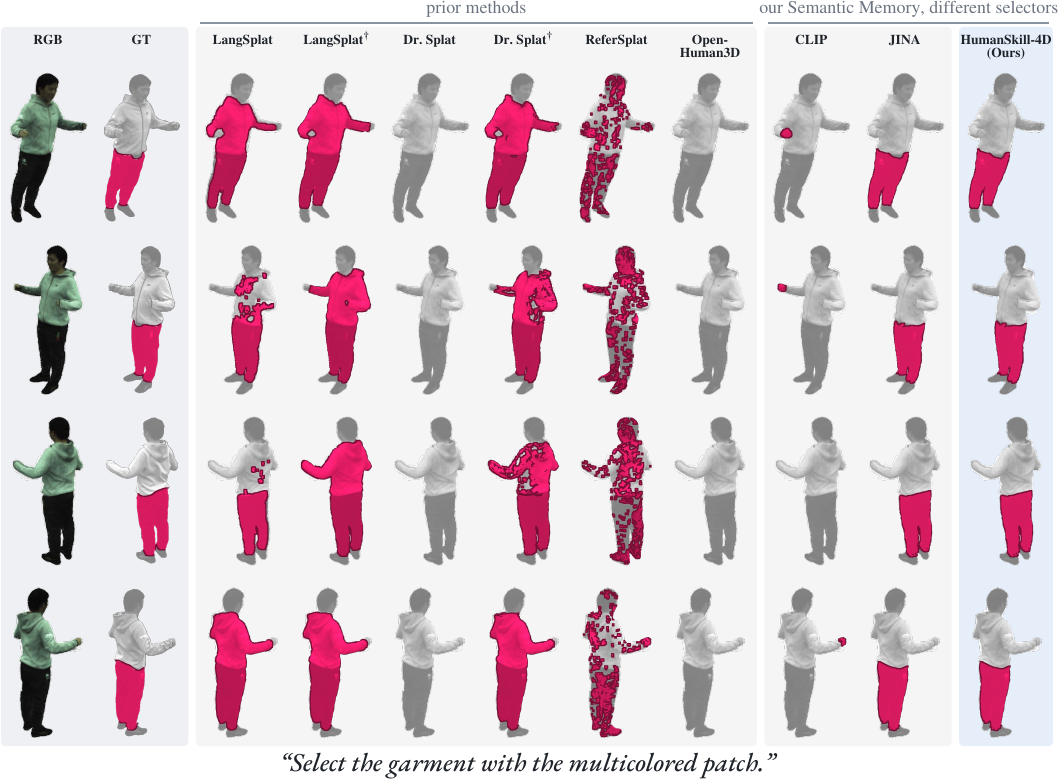}\par
\vspace{1.5pt}
\caption{\textbf{Qualitative 3D referring segmentation on \RefThreeD{}.} $^\dagger$~uses Sapiens2 masks.}
\label{fig:qual_comparison_rel}
\end{figure}

\textbf{3D referring segmentation.}\label{app:qual_comparison} Figs.~\ref{fig:qual_comparison} and~\ref{fig:qual_comparison_rel} show two queries from four held-out cameras. The CLIP and JINA columns keep our semantic memory but replace $\Phi_{\mathrm{MLLM}}$ with text-embedding retrieval, as in Tab.~\ref{tab:ablations} (middle) of the main paper. For \emph{``Select both hands''}, several methods find the hands, but ReferSplat covers most of the body, LangSplat also selects unrelated body regions, OpenHuman3D selects one shoe, CLIP adds small regions near the ankles, and JINA adds a forearm. For \emph{``Select the garment with the multicolored patch''}, prior predictions include unrelated clothing or body regions and CLIP selects only a hand, whereas JINA and our method select the patched lower garment alone. Neither CLIP nor JINA selects the target alone in both queries, whereas our method does.

\begin{figure}[!ht]
\centering
\includegraphics[width=\linewidth]{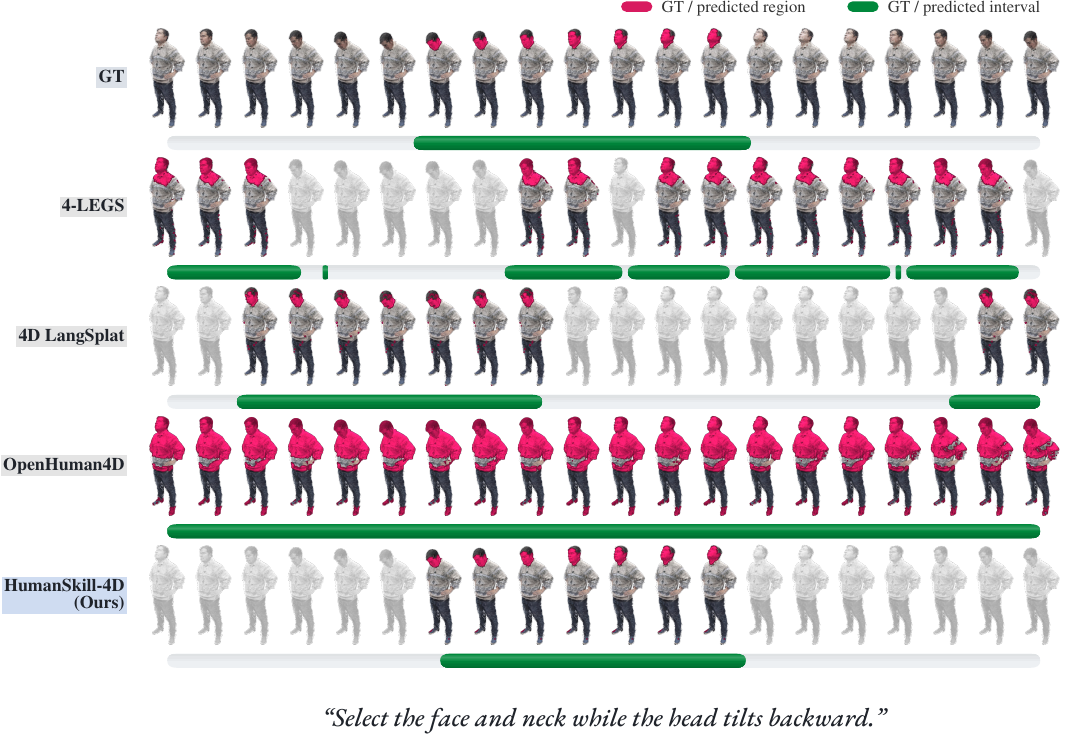}\par
\vspace{-.5em}
\caption{\textbf{Qualitative 4D referring segmentation on \RefHuMManFourD{}.}}
\label{fig:temporal_examples}
\end{figure}

\begin{figure}[!ht]
\centering
\includegraphics[width=\linewidth]{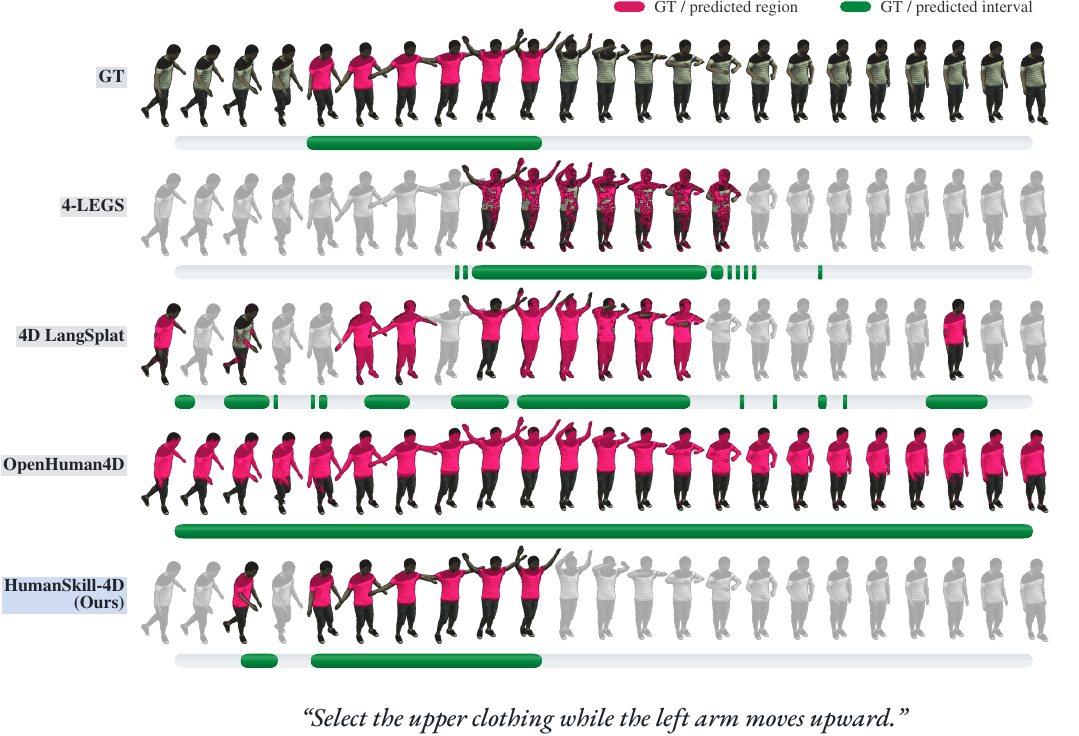}\par
\vspace{-.5em}
\caption{\textbf{Qualitative 4D referring segmentation on \RefZJUFourD{}.}}
\label{fig:temporal_z13}
\end{figure}

\textbf{4D referring segmentation.}\label{app:qual_temporal} Figs.~\ref{fig:temporal_examples}--\ref{fig:temporal_z13} show four queries, each from a held-out camera. For \emph{``Select the face and neck while the head tilts backward''} and \emph{``Select the right hand while the left foot is off the floor''}, our intervals closely match GT. For \emph{``Select the upper clothing while the left upper arm is nearly parallel to the ground''}, we recover both GT occurrences but split the first into two intervals. For \emph{``Select the upper clothing while the left arm moves upward''}, we return a short extra interval before the main lift. In contrast, 4-LEGS and 4D LangSplat often turn on outside GT intervals, and OpenHuman4D stays active throughout.

\begin{figure}[!ht]
\centering
\includegraphics[width=\linewidth]{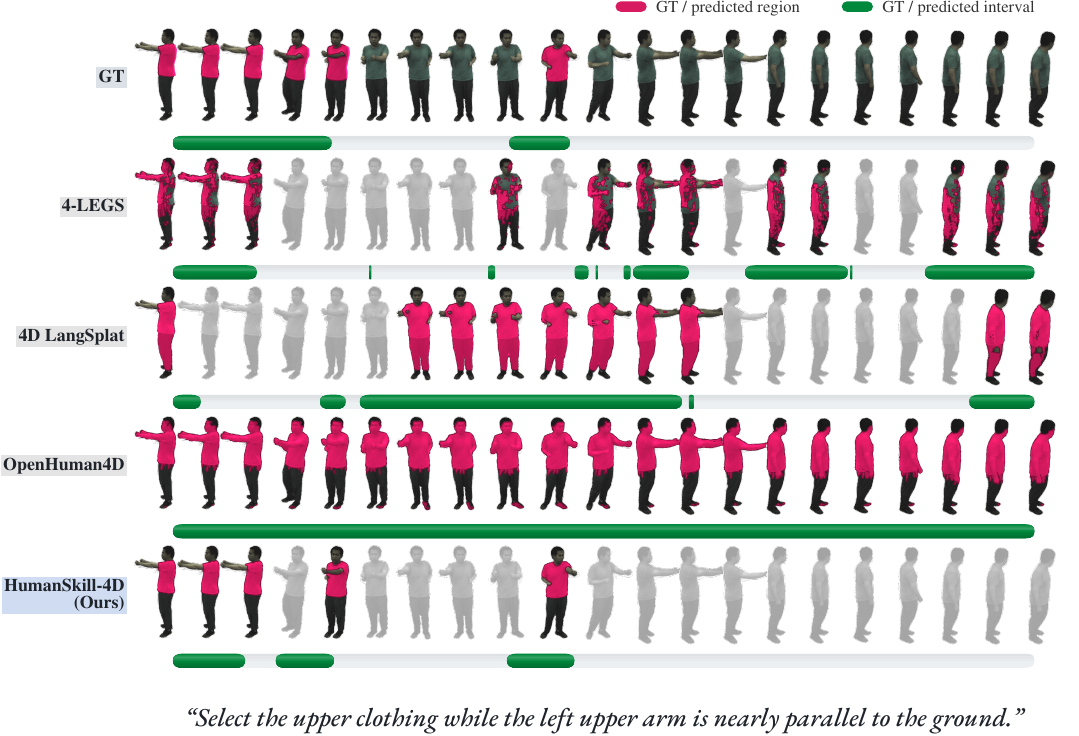}\par
\vspace{-.5em}
\caption{\textbf{Qualitative 4D referring segmentation on \RefZJUFourD{}.}}
\end{figure}

\begin{figure}[!ht]
\centering
\includegraphics[width=\linewidth]{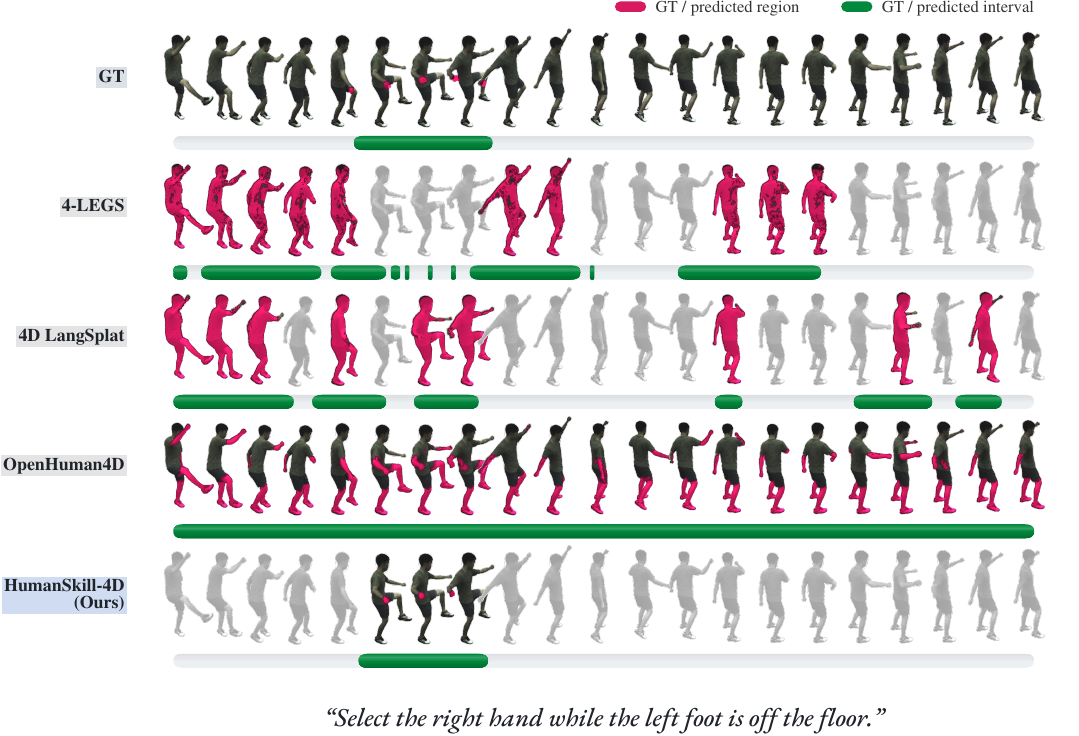}\par
\vspace{-.5em}
\caption{\textbf{Qualitative 4D referring segmentation on \RefZJUFourD{}.}}
\end{figure}

\section{Limitations and future directions}
\label{app:limits}

Because \Ours{} reuses a frozen parser and a shared registry, it needs no per-avatar training, but it selects regions only among the categories of $\Phi_{\mathrm{parse}}$ and checks motion only with the templates in $\mathcal{R}$. Details outside these categories, such as a logo or a watch, may therefore not be selectable, and motions outside these templates, such as dancing or carrying an object, may not be checked. Future work can extend both components within the same framework. For regions, a finer segmenter such as SAM~3~\citep{carion2026sam3}, which segments objects named by short noun phrases, could be used alongside $\Phi_{\mathrm{parse}}$ to register such details as categories. For motion, actions could be decomposed into part-level conditions, as in prior work on fine-grained human motion~\citep{delmas2022posescript,yazdian2025motionscript,zhang2023finemogen,wu2025finemotion}, or object-relative templates could be added given object trajectories.

\FloatBarrier

\section{Complete query list}
\label{app:query_list}

Tab.~\ref{tab:query_catalog_3d} lists the \RefThreeD{} queries of Tab.~\ref{tab:static3d} of the main paper, and Tabs.~\ref{tab:query_catalog_humman4d} and~\ref{tab:query_catalog_zju4d} list the \RefHuMManFourD{} and \RefZJUFourD{} query subsets of Tab.~\ref{tab:joint4d}.
\input{tables/appendix/reference_query_catalog_3d}
\input{tables/appendix/reference_query_catalog_humman4d}
\input{tables/appendix/reference_query_catalog_zju4d}

\section{MLLM instructions}
\label{app:prompts}

Tabs.~\ref{tab:prompt_describer}--\ref{tab:prompt_compiler} show the instructions and ICL examples given to $\Phi_{\mathrm{MLLM}}$.

\input{tables/appendix/reference_llm_prompts}

%% file: tables/appendix/reference_registry.tex
\begin{table}[t]
\vspace*{-\abovecaptionskip}
\caption{\textbf{Operator registry $\mathcal{R}$.} Templates grouped by measurement; counts in parentheses.}
\label{tab:registry}
\centering
\setlength{\tabcolsep}{3pt}
\begin{tabular}{@{}>{\raggedright\arraybackslash}p{1.95cm}>{\raggedright\arraybackslash}p{3.0cm}>{\raggedright\arraybackslash}p{3.15cm}>{\raggedright\arraybackslash}p{3.3cm}>{\raggedright\arraybackslash}p{1.55cm}@{}}
\toprule
Measurement type & Definition & State templates (20) & Change templates (26) & Coordinate frame \\
\midrule
Joint angle & Angle at a 3-joint hinge (e.g., shoulder--elbow--wrist) & (7) straight, broad bent, and 5 mutually exclusive bent degrees (slight, partial, right angle, deep, full) & (2) bend; straighten & Root-local \\
\midrule
Relative position & Point relative to a reference (state) or segment endpoint relative to its start (change), along a fixed-frame axis & (6) above/below, front/behind, left/right & (8) move up/down, forward/backward, left/right, inward/outward & Fixed \\
\midrule
Displacement & Point or body-root travel along a fixed-frame axis & (0) none & (8) up/down, forward/backward, left/right, inward/outward & Fixed \\
\midrule
Root rotation & Whole-body pitch and yaw & (1) neutral pitch & (4) lean forward\slash backward; turn clockwise\slash counterclockwise & Fixed \\
\midrule
Segment tilt & Angle of a 2-joint segment relative to gravity & (2) vertical; horizontal & (2) toward vertical; toward horizontal & Gravity \\
\midrule
Distance & Euclidean distance between 2 body points & (2) near; far & (2) approach; separate & Root-local \\
\midrule
Height above lowest joint & Height of a foot or ankle above the lowest of 24 body joints & (2) near the lowest joint; above the lowest joint & (0) none & Gravity \\
\bottomrule
\end{tabular}
\vspace{-8pt}
\end{table}

%% file: figures/appendix/running_templates.tex
\begin{wrapfigure}{r}{.3\linewidth}
\vspace{-\intextsep}
\setlength{\abovecaptionskip}{3pt}
\centering
\definecolor{regdark}{HTML}{8A5A00}
\begin{tikzpicture}[x=1cm,y=1cm]
\node[anchor=south west,inner sep=0] at (0,0) {\includegraphics[height=2.2cm]{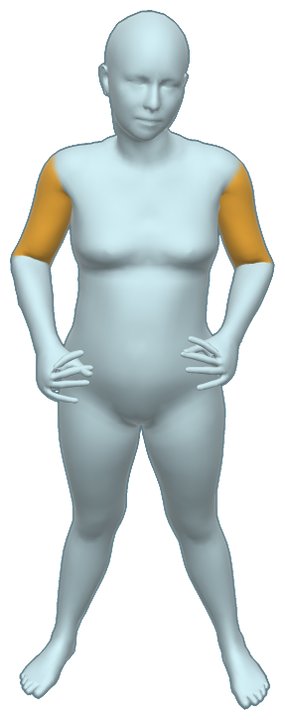}};
\draw[regdark,-latex,line width=.6pt] (0.836,1.371) -- (1.096,1.371);
\draw[regdark,-latex,line width=.6pt] (0.033,1.389) -- (-0.227,1.389);
\end{tikzpicture}
\hspace{0.5cm}\begin{tikzpicture}[x=1cm,y=1cm]
\node[anchor=south west,inner sep=0] at (0,0) {\includegraphics[height=2.2cm]{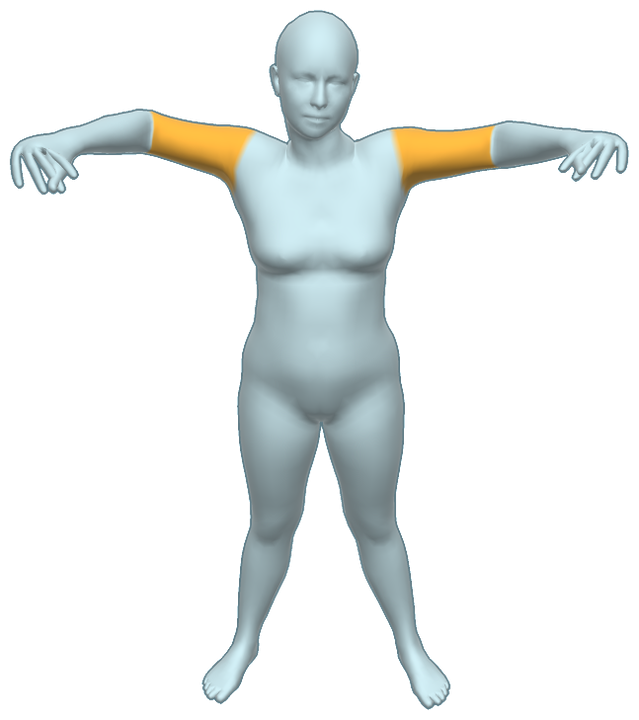}};
\draw[regdark,densely dashed,line width=.5pt] (1.140,1.720) -- (1.752,1.764);
\draw[regdark,densely dashed,line width=.5pt] (0.820,1.718) -- (0.231,1.838);
\end{tikzpicture}
\caption{\raggedright\textbf{Running-query templates.} \emph{Move outward} (left) and \emph{horizontal} (right) on both upper arms.}
\label{fig:f01_templates}
\end{wrapfigure}

%% file: figures/appendix/running_measurements.tex
\begin{figure}[!ht]
\centering
\begin{tikzpicture}[x=0.86mm,y=1mm,font=\footnotesize]
\definecolor{opamber}{HTML}{D08400}
\node[anchor=east] at (-6,7.5) {$\rho_1,\rho_2$: elbow offset (cm)};
\node[anchor=east] at (-6,-13.5) {$\rho_3,\rho_4$: upper-arm tilt ($^\circ$)};
\draw (-0.8,0) -- (0,0) node[pos=0,anchor=east,font=\scriptsize] {0};
\draw (-0.8,15) -- (0,15) node[pos=0,anchor=east,font=\scriptsize] {30};
\draw (-0.8,-21) -- (0,-21) node[pos=0,anchor=east,font=\scriptsize] {0};
\draw (-0.8,-6) -- (0,-6) node[pos=0,anchor=east,font=\scriptsize] {90};
\draw (0,0) -- (0,15); \draw (0,-21) -- (0,-6);
\draw[densely dotted,black!55] (0,-7.67) -- (103,-7.67) node[right,font=\scriptsize] {80};
\draw[black!35,line width=.5pt] (0,13.03) -- (1,13.04) -- (2,13.05) -- (3,13.05) -- (4,13.04) -- (5,13.03) -- (6,13.03) -- (7,13.02) -- (8,13.01) -- (9,13.02) -- (10,13.03) -- (11,13.04) -- (12,13.06) -- (13,13.07) -- (14,13.07) -- (15,13.04) -- (16,12.98) -- (17,12.88) -- (18,12.71) -- (19,12.47) -- (20,12.12) -- (21,11.67) -- (22,11.09) -- (23,10.4) -- (24,9.59) -- (25,8.72) -- (26,7.82) -- (27,6.92) -- (28,6.06) -- (29,5.27) -- (30,4.54) -- (31,3.92) -- (32,3.45) -- (33,3.13) -- (34,2.94) -- (35,2.84) -- (36,2.85) -- (37,2.95) -- (38,3.12) -- (39,3.33) -- (40,3.61) -- (41,3.99) -- (42,4.5) -- (43,5.15) -- (44,5.97) -- (45,6.89) -- (46,7.85) -- (47,8.8) -- (48,9.7) -- (49,10.53) -- (50,11.23) -- (51,11.79) -- (52,12.21) -- (53,12.51) -- (54,12.71) -- (55,12.84) -- (56,12.9) -- (57,12.92) -- (58,12.92) -- (59,12.92) -- (60,12.91) -- (61,12.89) -- (62,12.88) -- (63,12.88) -- (64,12.88) -- (65,12.89) -- (66,12.91) -- (67,12.94) -- (68,12.98) -- (69,13) -- (70,13) -- (71,12.99) -- (72,12.94) -- (73,12.84) -- (74,12.67) -- (75,12.42) -- (76,12.08) -- (77,11.64) -- (78,11.1) -- (79,10.48) -- (80,9.77) -- (81,9) -- (82,8.22) -- (83,7.42) -- (84,6.64) -- (85,5.89) -- (86,5.18) -- (87,4.56) -- (88,4.01) -- (89,3.55) -- (90,3.17) -- (91,2.87) -- (92,2.66) -- (93,2.54) -- (94,2.48) -- (95,2.49) -- (96,2.54) -- (97,2.65) -- (98,2.82) -- (99,3.01) -- (100,3.21) -- (101,3.41) -- (102,3.54);
\draw[opamber,line width=1.4pt] (35,2.84) -- (36,2.85) -- (37,2.95) -- (38,3.12) -- (39,3.33) -- (40,3.61) -- (41,3.99) -- (42,4.5) -- (43,5.15) -- (44,5.97) -- (45,6.89) -- (46,7.85) -- (47,8.8) -- (48,9.7) -- (49,10.53) -- (50,11.23) -- (51,11.79) -- (52,12.21) -- (53,12.51) -- (54,12.71) -- (55,12.84) -- (56,12.9) -- (57,12.92) -- (58,12.92);
\draw[opamber,line width=1.4pt] (94,2.48) -- (95,2.49) -- (96,2.54) -- (97,2.65) -- (98,2.82) -- (99,3.01) -- (100,3.21) -- (101,3.41) -- (102,3.54);
\draw[black!35,line width=.5pt,densely dashed] (0,12.65) -- (1,12.63) -- (2,12.59) -- (3,12.55) -- (4,12.49) -- (5,12.44) -- (6,12.39) -- (7,12.35) -- (8,12.32) -- (9,12.31) -- (10,12.31) -- (11,12.33) -- (12,12.36) -- (13,12.38) -- (14,12.4) -- (15,12.39) -- (16,12.36) -- (17,12.28) -- (18,12.15) -- (19,11.94) -- (20,11.62) -- (21,11.2) -- (22,10.66) -- (23,10.05) -- (24,9.35) -- (25,8.57) -- (26,7.75) -- (27,6.93) -- (28,6.12) -- (29,5.38) -- (30,4.77) -- (31,4.31) -- (32,3.94) -- (33,3.68) -- (34,3.53) -- (35,3.47) -- (36,3.45) -- (37,3.45) -- (38,3.49) -- (39,3.65) -- (40,3.89) -- (41,4.24) -- (42,4.69) -- (43,5.26) -- (44,5.96) -- (45,6.77) -- (46,7.67) -- (47,8.61) -- (48,9.53) -- (49,10.36) -- (50,11.08) -- (51,11.66) -- (52,12.1) -- (53,12.4) -- (54,12.59) -- (55,12.7) -- (56,12.73) -- (57,12.73) -- (58,12.7) -- (59,12.65) -- (60,12.6) -- (61,12.56) -- (62,12.53) -- (63,12.51) -- (64,12.51) -- (65,12.52) -- (66,12.55) -- (67,12.58) -- (68,12.63) -- (69,12.67) -- (70,12.69) -- (71,12.68) -- (72,12.63) -- (73,12.51) -- (74,12.31) -- (75,11.99) -- (76,11.58) -- (77,11.06) -- (78,10.46) -- (79,9.77) -- (80,9.04) -- (81,8.28) -- (82,7.5) -- (83,6.74) -- (84,6.02) -- (85,5.34) -- (86,4.73) -- (87,4.21) -- (88,3.8) -- (89,3.46) -- (90,3.21) -- (91,3.02) -- (92,2.88) -- (93,2.78) -- (94,2.71) -- (95,2.66) -- (96,2.67) -- (97,2.71) -- (98,2.79) -- (99,2.94) -- (100,3.12) -- (101,3.32) -- (102,3.45);
\draw[opamber,line width=1.4pt,densely dashed] (37,3.45) -- (38,3.49) -- (39,3.65) -- (40,3.89) -- (41,4.24) -- (42,4.69) -- (43,5.26) -- (44,5.96) -- (45,6.77) -- (46,7.67) -- (47,8.61) -- (48,9.53) -- (49,10.36) -- (50,11.08) -- (51,11.66) -- (52,12.1) -- (53,12.4) -- (54,12.59) -- (55,12.7) -- (56,12.73);
\draw[opamber,line width=1.4pt,densely dashed] (95,2.66) -- (96,2.67) -- (97,2.71) -- (98,2.79) -- (99,2.94) -- (100,3.12) -- (101,3.32) -- (102,3.45);
\draw[black!35,line width=.5pt] (0,-6.05) -- (1,-6.12) -- (2,-6.24) -- (3,-6.39) -- (4,-6.53) -- (5,-6.64) -- (6,-6.74) -- (7,-6.83) -- (8,-6.86) -- (9,-6.82) -- (10,-6.75) -- (11,-6.6) -- (12,-6.41) -- (13,-6.14) -- (14,-6.21) -- (15,-6.63) -- (16,-7.11) -- (17,-7.64) -- (18,-8.24) -- (19,-8.91) -- (20,-9.65) -- (21,-10.46) -- (22,-11.31) -- (23,-12.2) -- (24,-13.1) -- (25,-13.97) -- (26,-14.78) -- (27,-15.53) -- (28,-16.18) -- (29,-16.73) -- (30,-17.17) -- (31,-17.51) -- (32,-17.72) -- (33,-17.84) -- (34,-17.9) -- (35,-17.93) -- (36,-17.92) -- (37,-17.89) -- (38,-17.84) -- (39,-17.79) -- (40,-17.72) -- (41,-17.6) -- (42,-17.37) -- (43,-16.99) -- (44,-16.42) -- (45,-15.69) -- (46,-14.85) -- (47,-13.92) -- (48,-12.95) -- (49,-11.93) -- (50,-10.94) -- (51,-10) -- (52,-9.14) -- (53,-8.36) -- (54,-7.67) -- (55,-7.06) -- (56,-6.54) -- (57,-6.12) -- (58,-6.23) -- (59,-6.47) -- (60,-6.7) -- (61,-6.85) -- (62,-6.94) -- (63,-6.97) -- (64,-6.96) -- (65,-6.87) -- (66,-6.75) -- (67,-6.56) -- (68,-6.33) -- (69,-6.06) -- (70,-6.25) -- (71,-6.63) -- (72,-7.11) -- (73,-7.67) -- (74,-8.31) -- (75,-9) -- (76,-9.73) -- (77,-10.51) -- (78,-11.3) -- (79,-12.08) -- (80,-12.86) -- (81,-13.6) -- (82,-14.29) -- (83,-14.91) -- (84,-15.47) -- (85,-15.95) -- (86,-16.38) -- (87,-16.75) -- (88,-17.06) -- (89,-17.3) -- (90,-17.51) -- (91,-17.68) -- (92,-17.81) -- (93,-17.9) -- (94,-17.96) -- (95,-18) -- (96,-18.02) -- (97,-18.02) -- (98,-18.01) -- (99,-18.01) -- (100,-17.99) -- (101,-17.96) -- (102,-17.94);
\draw[opamber,line width=1.4pt] (0,-6.05) -- (1,-6.12) -- (2,-6.24) -- (3,-6.39) -- (4,-6.53) -- (5,-6.64) -- (6,-6.74) -- (7,-6.83) -- (8,-6.86) -- (9,-6.82) -- (10,-6.75) -- (11,-6.6) -- (12,-6.41) -- (13,-6.14) -- (14,-6.21) -- (15,-6.63) -- (16,-7.11) -- (17,-7.64);
\draw[opamber,line width=1.4pt] (55,-7.06) -- (56,-6.54) -- (57,-6.12) -- (58,-6.23) -- (59,-6.47) -- (60,-6.7) -- (61,-6.85) -- (62,-6.94) -- (63,-6.97) -- (64,-6.96) -- (65,-6.87) -- (66,-6.75) -- (67,-6.56) -- (68,-6.33) -- (69,-6.06) -- (70,-6.25) -- (71,-6.63) -- (72,-7.11);
\draw[black!35,line width=.5pt,densely dashed] (0,-6.42) -- (1,-6.55) -- (2,-6.74) -- (3,-6.94) -- (4,-7.14) -- (5,-7.33) -- (6,-7.48) -- (7,-7.53) -- (8,-7.55) -- (9,-7.52) -- (10,-7.41) -- (11,-7.25) -- (12,-7.02) -- (13,-6.75) -- (14,-6.42) -- (15,-6.05) -- (16,-6.57) -- (17,-7.18) -- (18,-7.85) -- (19,-8.58) -- (20,-9.38) -- (21,-10.2) -- (22,-11.04) -- (23,-11.87) -- (24,-12.71) -- (25,-13.53) -- (26,-14.32) -- (27,-15.05) -- (28,-15.71) -- (29,-16.28) -- (30,-16.71) -- (31,-17.03) -- (32,-17.29) -- (33,-17.47) -- (34,-17.58) -- (35,-17.64) -- (36,-17.68) -- (37,-17.72) -- (38,-17.76) -- (39,-17.72) -- (40,-17.61) -- (41,-17.42) -- (42,-17.14) -- (43,-16.75) -- (44,-16.25) -- (45,-15.59) -- (46,-14.81) -- (47,-13.91) -- (48,-12.94) -- (49,-11.94) -- (50,-10.94) -- (51,-9.97) -- (52,-9.06) -- (53,-8.22) -- (54,-7.46) -- (55,-6.79) -- (56,-6.21) -- (57,-6.29) -- (58,-6.7) -- (59,-7.06) -- (60,-7.32) -- (61,-7.53) -- (62,-7.67) -- (63,-7.76) -- (64,-7.78) -- (65,-7.73) -- (66,-7.6) -- (67,-7.4) -- (68,-7.11) -- (69,-6.74) -- (70,-6.29) -- (71,-6.23) -- (72,-6.82) -- (73,-7.5) -- (74,-8.22) -- (75,-9.02) -- (76,-9.85) -- (77,-10.69) -- (78,-11.53) -- (79,-12.35) -- (80,-13.12) -- (81,-13.84) -- (82,-14.52) -- (83,-15.13) -- (84,-15.67) -- (85,-16.16) -- (86,-16.56) -- (87,-16.9) -- (88,-17.16) -- (89,-17.38) -- (90,-17.56) -- (91,-17.7) -- (92,-17.82) -- (93,-17.91) -- (94,-17.97) -- (95,-18.03) -- (96,-18.08) -- (97,-18.13) -- (98,-18.15) -- (99,-18.14) -- (100,-18.1) -- (101,-18.04) -- (102,-18);
\draw[opamber,line width=1.4pt,densely dashed] (0,-6.42) -- (1,-6.55) -- (2,-6.74) -- (3,-6.94) -- (4,-7.14) -- (5,-7.33) -- (6,-7.48) -- (7,-7.53) -- (8,-7.55) -- (9,-7.52) -- (10,-7.41) -- (11,-7.25) -- (12,-7.02) -- (13,-6.75) -- (14,-6.42) -- (15,-6.05) -- (16,-6.57) -- (17,-7.18);
\draw[opamber,line width=1.4pt,densely dashed] (54,-7.46) -- (55,-6.79) -- (56,-6.21) -- (57,-6.29) -- (58,-6.7) -- (59,-7.06) -- (60,-7.32) -- (61,-7.53);
\draw[opamber,line width=1.4pt,densely dashed] (66,-7.6) -- (67,-7.4) -- (68,-7.11) -- (69,-6.74) -- (70,-6.29) -- (71,-6.23) -- (72,-6.82) -- (73,-7.5);
\draw (0,-23.5) -- (103,-23.5);
\foreach \t in {0,20,40,60,80,100}{\draw (\t,-23.5) -- (\t,-24.2) node[below=-1pt] {\t};}
\node[anchor=north] at (51.5,-27.7) {Frame};
\end{tikzpicture}
\caption{\textbf{Operator measurements of the running query.} The top panel shows how far each elbow lies outward from its shoulder ($\rho_1,\rho_2$), and the bottom panel shows the angle of each upper arm from vertical ($\rho_3,\rho_4$). Solid lines are the left arm and dashed lines the right, and amber marks the frames each operator returns.}
\label{fig:f01_measure}
\end{figure}

%% file: tables/appendix/reference_relations.tex
\providecommand{\promptsep}{\arrayrulecolor{black!22}\midrule\arrayrulecolor{black}}
\begin{table}[!ht]
\vspace*{-\abovecaptionskip}
\caption{\textbf{Temporal relations.}}
\label{tab:relations}
\centering
\setlength{\tabcolsep}{2pt}
\begin{tabular}{@{}p{2.4cm}p{3.5cm}p{\dimexpr\linewidth-5.9cm-8pt\relax}@{}}
\toprule
Relation & Inputs & Output \\
\midrule
CoOccurrence & 2 or more movements & occurrences formed by intervals shared by every movement \\
\promptsep
AllStates & 2 or more states & frames shared by all states \\
\promptsep
Sequence & movements in order; optional leading state & span from the first to the last movement, linking each last occurrence to its nearest eligible predecessors with no maximum gap; consecutive movements may share a turning frame; a leading state must end before the first movement starts and does not extend the span \\
\promptsep
Alternation & 2 movements A and B & spans of alternating chains of 3 or more occurrences; consecutive movements have disjoint interiors and may share a turning frame \\
\promptsep
StateAtEnd & a movement and a state & occurrences whose last frame lies in the state \\
\promptsep
StateThroughout & a movement and a state & occurrences fully covered by the state \\
\promptsep
During & a movement and a state & each occurrence clipped to every overlapping state interval \\
\bottomrule
\end{tabular}
\end{table}

%% file: figures/appendix/f01_trace.tex
\begin{figure}[!ht]
\centering
\begin{tikzpicture}[x=0.86mm,y=4.6mm,font=\footnotesize]
\newcommand{\bars}[3]{\foreach \a/\b in {#2}{\fill[#3] (\a,-#1+0.18) rectangle (\b+1,-#1-0.18);}}
\foreach \r/\lab in {1/{$\rho_1$: left upper arm moves outward},2/{$\rho_2$: right upper arm moves outward},3/{CoOccurrence$(\rho_1,\rho_2)$},4/{$\rho_3$: left upper arm horizontal},5/{$\rho_4$: right upper arm horizontal},6/{AllStates$(\rho_3,\rho_4)$},7/{StateAtEnd}}{\node[anchor=east] at (-1.5,-\r) {\lab};\draw[black!15] (0,-\r) -- (103,-\r);}
\bars{1}{35/58,94/102}{black!35}
\bars{2}{37/56,95/102}{black!35}
\bars{3}{37/56,95/102}{black!65}
\bars{4}{0/17,55/72}{black!35}
\bars{5}{0/17,54/61,66/73}{black!35}
\bars{6}{0/17,55/61,66/72}{black!65}
\bars{7}{37/56}{whenblue}
\foreach \t in {56,102}{\draw[densely dashed,black!70] (\t+0.5,-0.45) -- (\t+0.5,-7.4);}
\node[anchor=south,inner sep=0] at (56.5,-0.45) {\includegraphics[height=1.5cm]{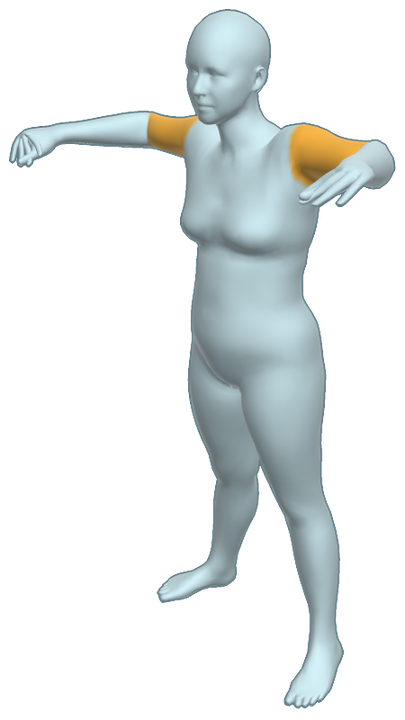}};
\node[anchor=south,inner sep=0] at (102.5,-0.45) {\includegraphics[height=1.5cm]{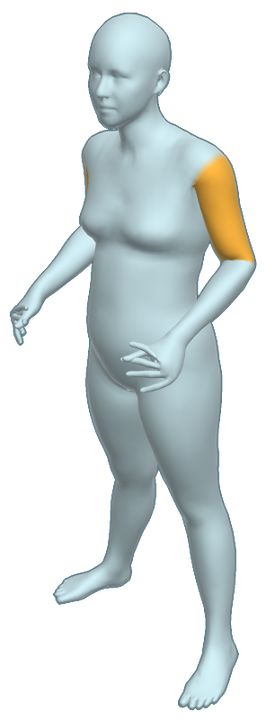}};
\draw (0,-7.75) -- (103,-7.75);
\foreach \t in {0,20,40,60,80,100}{\draw (\t,-7.75) -- (\t,-7.9) node[below=-1pt] {\t};}
\node[anchor=north] at (51.5,-8.45) {Frame};
\end{tikzpicture}
\caption{\textbf{Execution record of the running query.} Bars mark the frames returned by each operator and relation; dashed lines and bodies mark the last frame of each co-occurrence.}
\label{fig:f01_trace}
\end{figure}

%% file: tables/appendix/reference_llm_cards.tex
\providecommand{\promptsep}{\arrayrulecolor{black!22}\midrule\arrayrulecolor{black}}
\begin{table}[!ht]
\vspace*{-\abovecaptionskip}
\caption{\textbf{Text cards used for the running query.}}
\label{tab:llm_cards}
\centering
\setlength{\tabcolsep}{4pt}
\begin{tabular}{@{}p{4.2cm}p{\dimexpr\linewidth-4.2cm-8pt\relax}@{}}
\toprule
Card (kind) & Card summary \\
\midrule
\emph{move outward} (change) & \emph{Measures:} the segment endpoint, relative to its start, moves toward its anatomical outer side. \emph{Parts:} arm, forearm, leg, lower leg, thigh, upper arm. \\
\promptsep
\emph{horizontal} (state) & \emph{Measures:} the segment is horizontal relative to gravity. \\
\midrule
\textit{CoOccurrence} (relation) & \emph{Inputs:} two or more movements. \emph{Meaning:} the movements must hold on the same frames. \emph{Returns:} every shared portion on which all of the movements hold. \\
\promptsep
\textit{AllStates} (relation) & \emph{Inputs:} two or more states. \emph{Meaning:} the states must hold on the same frames. \emph{Returns:} the frames on which every state holds. \\
\promptsep
\textit{StateAtEnd} (relation) & \emph{Inputs:} a movement, then a state. \emph{Meaning:} the state describes the pose in which the movement ends. \emph{Returns:} the movement occurrence whose last frame lies inside the state. \\
\bottomrule
\end{tabular}
\vspace{-8pt}
\end{table}

%% file: tables/appendix/reference_method_components.tex
\begin{table}[!t]
\vspace*{-\abovecaptionskip}
\caption{\textbf{Method components} for the computational cost comparison in Tab.~\ref{tab:cost}. Stored information lists component types. Segmentation: 2D mask source; $\times$: none.}
\label{tab:components}
\centering
\setlength{\tabcolsep}{6pt}
\setlength{\abovetopsep}{4pt}
\setlength{\aboverulesep}{0pt}
\setlength{\belowrulesep}{0pt}
\setlength{\extrarowheight}{1.5pt}
\renewcommand{\arraystretch}{1.08}
\begin{tabular}{@{}l!{\color{black!25}\vrule}l!{\color{black!25}\vrule}l!{\color{black!25}\vrule}l@{}}
\toprule
Method & Segmentation & Interpretation & Stored information \\
\midrule
\rowcolor{grouprow}\multicolumn{4}{@{}l}{3D referring segmentation} \\
LangSplat & SAM & CLIP & Features + networks \\
ReferSplat & Grounded-SAM & BERT + attention & Features + networks \\
OpenHuman3D & SAM & HumanCLIP & Masks + features \\
\textbf{\Ours{}} & Sapiens2 & MLLM & Scores + text \\
\midrule
\rowcolor{grouprow}\multicolumn{4}{@{}l}{4D referring segmentation} \\
4-LEGS & $\times$ & ViCLIP & Features + networks \\
4D LangSplat & SAM + DEVA & CLIP + E5 & Features + networks \\
OpenHuman4D & SAM/SAM~2 & HumanCLIP & Masks + features \\
\textbf{\Ours{}} & Sapiens2 & MLLM + operators & Scores + text + motion \\
\bottomrule
\end{tabular}
\vspace{-8pt}
\end{table}

%% file: tables/appendix/ablation_llm_small.tex
\caption{\textbf{Smaller MLLMs.} 3D mIoU and 4D vIoU (\%).}
\label{tab:llm_small}
\centering
\setlength{\tabcolsep}{3pt}
\begin{tabular}{@{}lcc@{}}
\toprule
Model & mIoU & vIoU \\
\midrule
Gemma 4 E2B & 60.2 & 0.0 \\
Ministral 3 8B & 63.1 & 20.4 \\
Qwen3-VL-8B & 78.4 & 16.2 \\
Phi-4 & 49.9 & 2.6 \\
\bottomrule
\end{tabular}

%% file: tables/appendix/reference_query_catalog_3d.tex
\begin{longtable}{@{}>{}l>{\raggedright\arraybackslash}p{0.92\linewidth}@{}}
\caption{\textbf{\RefThreeD{} queries.}}\label{tab:query_catalog_3d}\\
\toprule
ID & Query \\
\midrule
\endfirsthead
\toprule
ID & Query \\
\midrule
\endhead
\textbf{Q01} & Select both hands. \\
\textbf{Q02} & Select both arms. \\
\textbf{Q03} & Select both shoes. \\
\textbf{Q04} & Select the upper clothing and lower clothing. \\
\textbf{Q05} & Select the garment with a zipper down the front. \\
\addlinespace[2pt]
\textbf{Q06} & Select the yellow garment. \\
\textbf{Q07} & Select the garment with long sleeves. \\
\textbf{Q08} & Select the garment with the multicolored patch. \\
\textbf{Q09} & Select the sleeveless garment. \\
\textbf{Q10} & Select the denim garment. \\
\addlinespace[2pt]
\textbf{Q11} & Select the garment worn below the shirt. \\
\textbf{Q12} & Select the garment with a crown-shaped graphic on the front. \\
\textbf{Q13} & Select the garment with horizontal stripes across the front. \\
\textbf{Q14} & Select the garment with a large oval graphic on the front. \\
\textbf{Q15} & Select the garment with a white drawstring. \\
\addlinespace[2pt]
\textbf{Q16} & Select the accessory worn on the face. \\
\textbf{Q17} & Select the items with gold-colored crossing markings. \\
\textbf{Q18} & Select the garment with a hood and a large front pocket. \\
\textbf{Q19} & Select the garment with a plain upper section and horizontal stripes below it. \\
\textbf{Q20} & Select both body parts used to grasp objects. \\
\bottomrule
\vspace{-20pt}
\end{longtable}

%% file: tables/appendix/reference_query_catalog_humman4d.tex
\begin{longtable}{@{}>{}l>{\raggedright\arraybackslash}p{0.92\linewidth}@{}}
\caption{\textbf{\RefHuMManFourD{} queries.}}\label{tab:query_catalog_humman4d}\\
\toprule
ID & Query \\
\midrule
\endfirsthead
\textbf{F01} & Select the upper clothing while both upper arms lift outward until they are parallel to the ground. \\
\textbf{F02} & Select the left hand while the person lowers into a squat. \\
\textbf{F03} & Select the lower clothing while the left foot moves forward and the right foot remains on the ground. \\
\textbf{F04} & Select the lower clothing while the left knee bends. \\
\textbf{F05} & Select the right hand while the right knee bends and the body lowers. \\
\addlinespace[2pt]
\textbf{F06} & Select the lower clothing while the right knee lifts. \\
\textbf{F07} & Select the face and neck while the head tilts backward. \\
\textbf{F08} & Select the lower clothing while the left knee lifts. \\
\textbf{F09} & Select the lower clothing while the right knee straightens. \\
\textbf{F10} & Select the lower clothing while the body lowers. \\
\addlinespace[2pt]
\textbf{F11} & Select the lower clothing while the right foot moves to the right. \\
\textbf{F12} & Select the left hand while the left knee bends and the body lowers. \\
\textbf{F13} & Select the lower clothing while the left foot moves backward. \\
\textbf{F14} & Select the left hand while it moves toward the right knee as the right knee moves upward. \\
\textbf{F15} & Select the left hand while the left arm moves forward and rotates toward horizontal. \\
\addlinespace[2pt]
\textbf{F16} & Select the lower clothing while the left knee straightens. \\
\textbf{F17} & Select the lower clothing while the right knee bends and the body lowers. \\
\textbf{F18} & Select the left shoe while the left foot moves to the left. \\
\textbf{F19} & Select the upper clothing while the left upper arm rotates toward horizontal as the left foot moves to the left. \\
\textbf{F20} & Select the lower clothing while the left knee bends and the body lowers. \\
\bottomrule
\vspace{-20pt}
\end{longtable}

%% file: tables/appendix/reference_query_catalog_zju4d.tex
\begin{longtable}{@{}>{}l>{\raggedright\arraybackslash}p{0.92\linewidth}@{}}
\caption{\textbf{\RefZJUFourD{} queries.}}\label{tab:query_catalog_zju4d}\\
\toprule
ID & Query \\
\midrule
\endfirsthead
\toprule
ID & Query \\
\midrule
\endhead
\textbf{Z01} & Select the right leg each time the right knee bends. \\
\textbf{Z02} & Select the left leg while the left knee bends. \\
\textbf{Z03} & Select the upper clothing while the left upper arm is nearly parallel to the ground. \\
\textbf{Z04} & Select the right hand while the right upper arm is nearly parallel to the ground. \\
\textbf{Z05} & Select the right shoe while the whole right leg lifts until it is nearly parallel to the ground. \\
\addlinespace[2pt]
\textbf{Z06} & Select the left shoe while the left foot is off the floor. \\
\textbf{Z07} & Select the left hand while it is above the left shoulder. \\
\textbf{Z08} & Select the right shoe while the right thigh is nearly parallel to the ground. \\
\textbf{Z09} & Select the left shoe while the left thigh is nearly parallel to the ground. \\
\textbf{Z10} & Select the right shoe while the right thigh is nearly parallel to the ground. \\
\addlinespace[2pt]
\textbf{Z11} & Select the left hand while it is above the left shoulder. \\
\textbf{Z12} & Select the left hand while it is above the head. \\
\textbf{Z13} & Select the upper clothing while the left arm moves upward. \\
\textbf{Z14} & Select the left shoe while the whole left leg lifts until it is nearly parallel to the ground. \\
\textbf{Z15} & Select the right shoe while the right knee bends. \\
\addlinespace[2pt]
\textbf{Z16} & Select the right hand while it is above the head. \\
\textbf{Z17} & Select the left shoe while the left foot is off the floor. \\
\textbf{Z18} & Select the left hand while it is above the head. \\
\textbf{Z19} & Select the right shoe while the whole right leg lifts until it is nearly parallel to the ground. \\
\textbf{Z20} & Select the right hand while the left foot is off the floor. \\
\bottomrule
\end{longtable}

%% file: tables/appendix/reference_llm_prompts.tex
\providecommand{\promptsep}{\arrayrulecolor{black!22}\midrule\arrayrulecolor{black}}
\begin{table}[!ht]
{\small\frenchspacing
\centering
\setlength{\tabcolsep}{4pt}
\caption{\textbf{Instruction for the appearance describer.} [\dots] marks omissions.}\label{tab:prompt_describer}
\begin{tabular}{@{}p{1.75cm}p{\dimexpr\linewidth-1.75cm-8pt\relax}@{}}
\toprule
\textbf{Role} & You are the Appearance Describer. Record visible appearance for the supplied categories from synchronized images. \\
\promptsep
\textbf{Task} & [\dots] Describe every supplied category by combining compatible evidence from all supplied views. Preserve visible colors, graphics, readable text, logos, material, shape, texture, coverings, accessories, and feature placement. [\dots] \\
\promptsep
\textbf{Rules} & Your evidence is the supplied synchronized multiview RGB images. Treat them as one observation set and describe only appearance. [\dots] Combine complementary front, back, and side details [\dots] Left and right refer to the person's anatomical sides, not the image sides. \\
\promptsep
\textbf{Input} & \{category IDs and names\}; \{four views from the frame closest to a T-pose\} \\
\bottomrule
\end{tabular}
}
\end{table}

\begin{table}[!ht]
{\small\frenchspacing
\centering
\setlength{\tabcolsep}{4pt}
\caption{\textbf{Instruction and ICL example for the 3D selector.} [\dots] marks omissions.}\label{tab:prompt_selector}
\begin{tabular}{@{}p{1.75cm}p{\dimexpr\linewidth-1.75cm-8pt\relax}@{}}
\toprule
\textbf{Role} & You are the 3D Semantic Selector. Map the requested static target to categories supported by Semantic Memory. \\
\promptsep
\textbf{Task} & Call \texttt{select\_semantic\_labels} once with the stored categories that match the full static request, or abstain when no supported match exists. \\
\promptsep
\textbf{Rules} & Use category names, their ordinary meaning and synonyms for structural identity, and stored descriptions for appearance. [\dots] Every requested color, material, texture, logo, graphic, shape, or placement detail needs compatible stored evidence. [\dots] Select only categories present in the supplied inventory. \\
\promptsep
\textbf{Input} & \{query\}; \{inventory rows: category ID, name, description, status\} \\
\midrule
\textbf{ICL input} & \texttt{\{"query": "Find the dark shirt with a small front logo and a large rear graphic.", "semantic\_inventory": [\{"id": 1, "name": "upper\_clothing", "text": "A dark shirt has a small front logo and a large graphic across the back.", \dots\}, \{"id": 2, "name": "lower\_clothing", \dots\}]\}} \\
\promptsep
\textbf{ICL output} & \texttt{\{"activation\_label\_ids": [1], "abstain": false, "reason": "The stored upper-clothing description matches both requested graphics."\}} \\
\bottomrule
\end{tabular}
}
\end{table}

\makeatletter\setlength{\@fptop}{0pt}\makeatother
\begin{table}[!ht]
{\small\frenchspacing
\centering
\setlength{\tabcolsep}{4pt}
\caption{\textbf{Instruction and ICL example for the 4D compiler.} [\dots] marks omissions.}\label{tab:prompt_compiler}
\begin{tabular}{@{}p{1.75cm}p{\dimexpr\linewidth-1.75cm-8pt\relax}@{}}
\toprule
\textbf{Role} & [\dots] First decide whether the exact query is expressible using the supplied Semantic Memory and body measurements, then return its program or the specific missing support. [\dots] \\
\promptsep
\textbf{Task} & Return exactly one tool call. Use \texttt{submit\_component\_bindings} for a fully supported query: supply its answer target, target label IDs, motion or state components, temporal composition, and Registry bindings. Use \texttt{report\_query\_unsupported} when the supplied inventory or cards cannot express the full request, and identify the missing support. \\
\promptsep
\textbf{Rules} & The answer target and the subject of a WHEN condition are independent: the condition may describe any requested body part, including a part different from the selected target. [\dots] One component describes one movement or one state of a measurable part or bilateral pair. [\dots] A destination pose qualifies the complete movement that ends in it with \texttt{state\_at\_end}. [\dots] Report unsupported if the target lacks inventory support, a required anatomical side is missing, no body motion or state is requested, or any required measurement or temporal relation cannot be expressed. \\
\promptsep
\textbf{Input} & \{query\}; \{inventory\}; \{template and relation text cards as in Tab.~\ref{tab:llm_cards}\} \\
\midrule
\textbf{ICL input} & \texttt{\{"query": "Select the buckle while the body turns counterclockwise.", "semantic\_inventory": [\{"id": 29, "name": "buckle"\}]\}} \\
\promptsep
\textbf{ICL output} & \texttt{\{"target\_phrase": "the buckle", "components": [\{"component\_id": "c01", "kind": "motion", "phrase": "the body turns counterclockwise", \dots\}], "root": "c01", "target\_label\_ids": [29], "bindings\_by\_component": [\{\dots, "bindings": [\{"atom\_id": "root\_turn\_counterclockwise", "part": "body", \dots\}]\}], \dots\}} \\
\bottomrule
\end{tabular}
}
\end{table}